%% file: main.tex
\documentclass{article} 
\usepackage{iclr2025_conference,times}
\input{math_commands.tex}

\usepackage{hyperref}
\usepackage{url}
\usepackage{graphicx}
\usepackage{caption}
\usepackage{makecell}
\usepackage{colortbl}
\usepackage{xcolor}
\usepackage{multirow}
\usepackage{MnSymbol}
\usepackage{amsthm}
\definecolor{champagne}{rgb}{0.968, 0.906, 0.808}
\definecolor{palatinateblue}{rgb}{0.15, 0.23, 0.89}
\usepackage{booktabs} 
\usepackage{enumitem}
\usepackage{algorithm}
\usepackage{algpseudocode}
\usepackage{adjustbox}
\usepackage{fontawesome5}

\usepackage{acronym}
\newacro{CNN}[CNN]{convolutional neural network}
\newacro{SSM}[SSM]{Structured State-Space Model}
\newacro{RNN}[RNN]{Recurrent Neural Network}
\newacro{FACTS}[FACTS]{\tbf{FACT}ored \tbf{S}tate-space}
\newtheorem{theorem}{Theorem}
\newtheorem*{theorem*}{Theorem}

\DeclareMathOperator{\FACTS}{\mathbf{FACTS}}

\theoremstyle{definition}
\newtheorem{definition}{Definition}

\newcommand{\LPE}{L.P.E.}
\newcommand{\RPI}{R.P.I.}

\title{FACTS: A Factored State-Space Framework For World Modelling}

\author{Li Nanbo$^1$\thanks{Correspondence to \texttt{nanbo.li@kaust.edu.sa}}~, 
Firas Laakom$^1$, Yucheng Xu$^2$, Wenyi Wang$^1$, J\"{u}rgen Schmidhuber$^{1,3}$  \\
$^1$Center of Excellence for Generative AI, KAUST, Saudi Arabia \\
$^2$School of Informatics, University of Edinburgh, United Kingdom \\
$^3$The Swiss AI Lab, IDSIA, USI \& SUPSI, Switzerland
}

\newcommand{\tbf}[1]{\textbf{#1}}

\newcommand{\win}[1]{\textcolor{red}{\textbf{#1}}}    
\newcommand{\run}[1]{\textcolor{blue}{#1}}            

\iclrfinalcopy 
\begin{document}

\maketitle

\begin{abstract}
World modelling is essential for understanding and predicting the dynamics of complex systems by learning both spatial and temporal dependencies. However, current frameworks, such as Transformers and selective state-space models like Mambas, exhibit limitations in efficiently encoding spatial and temporal structures, particularly in scenarios requiring long-term high-dimensional sequence modelling. To address these issues, we propose a novel recurrent framework, the \tbf{FACT}ored \tbf{S}tate-space (\tbf{FACTS}) model, for spatial-temporal world modelling. The FACTS framework constructs a graph-structured memory with a routing mechanism that learns permutable memory representations, ensuring invariance to input permutations while adapting through selective state-space propagation. Furthermore, FACTS supports parallel computation of high-dimensional sequences. We empirically evaluate FACTS across diverse tasks, including multivariate time series forecasting, object-centric world modelling, and spatial-temporal graph prediction, demonstrating that it consistently outperforms or matches specialised state-of-the-art models, despite its general-purpose world modelling design. Code available at:~\faGithub~\textcolor{palatinateblue}{\url{https://github.com/NanboLi/FACTS}}.
\end{abstract}

\section{Introduction}
\label{intro}
World modelling~\citep{schmidhuber1990making,schmidhuber2015learning,ha2018world} aims to create an internal representation of the environment for an AI system, enabling it to represent~\citep{hafner2019learning,hafner2023mastering}, understand~\citep{schrittwieser2020mastering,hafner2020mastering}, and predict~\citep{ha2018world,micheli2022transformers} the dynamics of complex environments. This capability is crucial for various domains, including autonomous systems, robotics, and financial forecasting, where accurate predictions depend on effectively capturing both spatial and temporal dependencies~\citep{hafner2019learning,ha2018world}. Consequently, spatial-temporal learning~\citep{liu2023itransformer,hochreiter1997long,wu2022timesnet,oreshkin2019n} emerges as a key challenge in world modelling, as approaches must balance the complexities of modelling high-dimensional sequential data while maintaining robust long-term predictive power.

Despite significant advancements, current spatial-temporal learning frameworks, used in world modelling, based on  Transformers~\citep{schmidhuber1992fwp,vaswani2017attention,schlag2021linear} and RNNs~\citep{schmidhuber2015learning,ha2018world,hafner2019learning,hafner2020mastering} backbones, face limitations in fully capturing the complexities of high-dimensional spatial-temporal data. Transformer, though powerful~\citep{chen2022transdreamer,robine2023transformer,micheli2022transformers}, are inefficient for long-term tasks due to their quadratic scaling and limited context windows~\citep{zhang2022opt}. On the other hand, RNNs provide a more structured approach to sequential data. However, their efficacy is hindered by the vanishing gradients \citep{hochreiter1991untersuchungen,pascanu2013difficulty}. The primary challenges in spatial-temporal learning arise from the high dimensionality of the data and the necessity to preserve long-term dependencies~\citep{hochreiter2001gradient,tallec2018can}.

Recently, there has been a growing interest in Structured State-Space Models (SSMs) for world modelling~\citep{gu2023mamba,hafner2023mastering,samsami2024mastering} using latent state-space representations. These representations allow for the modelling of underlying dynamics, where latent states evolve over time according to the governing equations~\citep{wang2024state}. However, while SSMs have demonstrated improved capacity for capturing temporal dynamics~\citep{gu2023mamba,wang2024state,baron20232}, they often lack efficient mechanisms to handle high-dimensional spatial data. To address this limitation, recent approaches often impose rigid structural constraints on their state spaces, such as diagonal~\citep{gu2023mamba,gupta2022diagonal,gupta2022simplifying} or block-diagonal structures~\citep{dao2024transformers}, to capture invariant components throughout the sequence. This assumption, that specific dimensions of the state space correspond to consistent patterns over time, can be restrictive in world modelling scenarios where the relationship between state-space dimensions and input features evolves dynamically.

For example, in a dynamic system involving multiple agents (e.g. robots or sensors), where the positions of the agents change over time, to capture this dynamism, current SSMs require learning distinct representations for essentially identical scenarios at each time step as the agent locations change. This redundancy leads to an inefficient use of model capacity and data, ultimately limiting the model’s ability to effectively capture the dynamics of the interactions between inputs and states. Therefore, there is a need for a consistent dynamic mapping between inputs and latent states to enhance spatial-temporal modelling capabilities and enable more efficient history compression~\citep{schmidhuber1992learning,schmidhuber2003exploring}, which is essential for robust long-term prediction power. Another limitation of current SSMs is their inability to capture redundancy in the input space itself. In many cases, each agent's state (e.g., position, speed, direction) may contributes to the world’s overall understanding, but the identity or order of the agents do not matter, i.e., only their interactions are crucial for making accurate predictions.  In such instances, swapping agents should not alter the predictions and the world understanding. However, current SSM-based methods, typically based on linear transformations, fail to account for this and can perceive identical scenarios as different based on input order, hindering its ability to capture regularities and making them unsuitable for sequential modelling in various applications.
\begin{figure}[t]
    \begin{center}
    \includegraphics[width=0.75\linewidth]{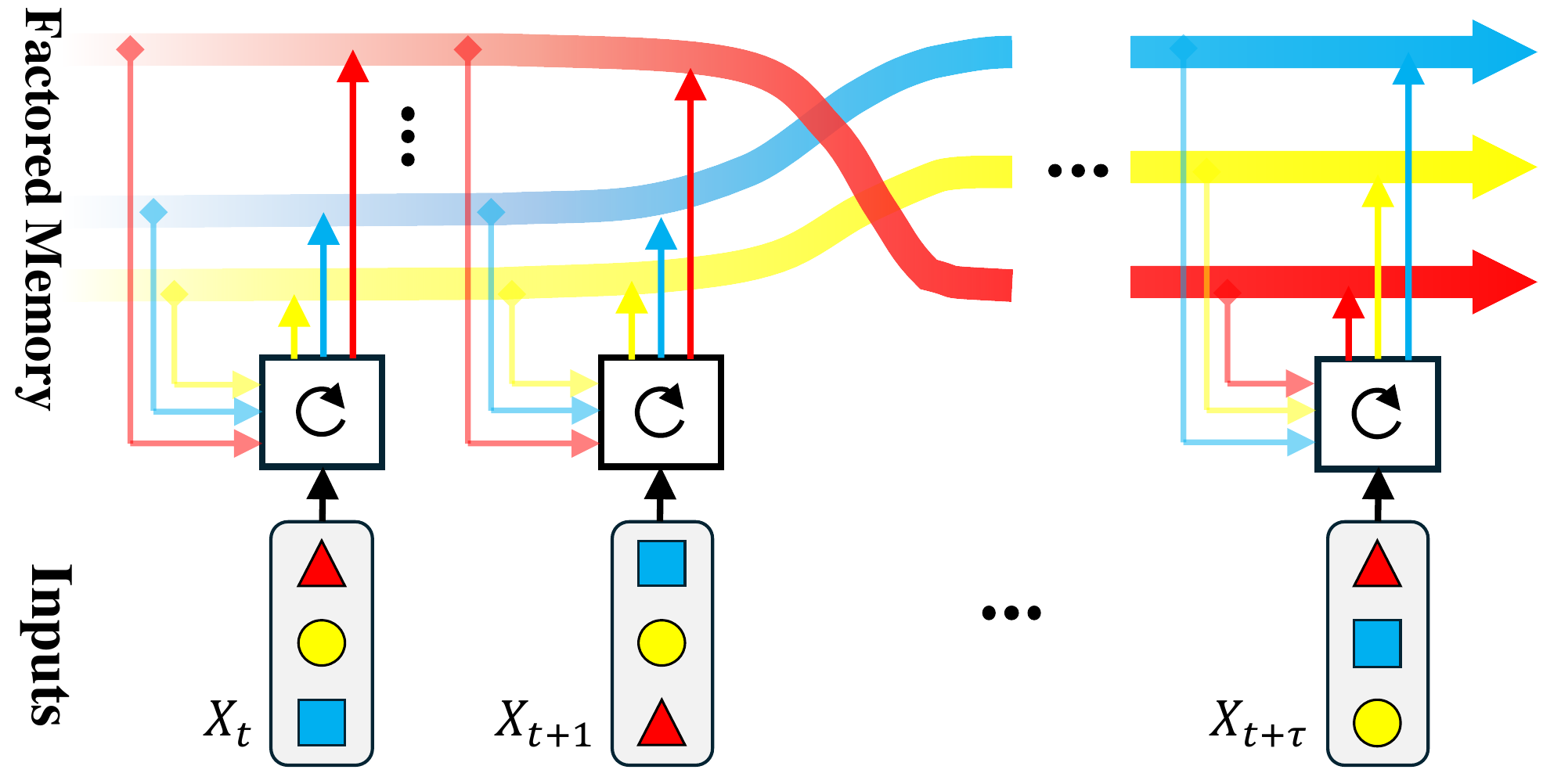} 
    \end{center}
    \caption{\textbf{Overview of \ac{FACTS} Architecture.} The FACTS framework constructs a factored state-space memory, allowing for flexible representations (e.g. graphs and sets). Sequential inputs (e.g. $X_t$) are processed through a selective memory-input interaction mechanism (denoted by the circular icon $\lcirclearrowright$), which determines how the inputs interact with and update factored memory. The different coloured pathways represent distinct latent factors, whose dynamics evolve over time based on these interactions. The design ensures that the memory update is permutation-invariant with respect to the input features, enabling FACTS to capture and track meaningful algorithmic regularities for accurate future predictions. }
    \label{fig:facts}
\vspace{-1em}
\end{figure}

To address these challenges, in this paper, we propose the \ac{FACTS} model, a novel recurrent framework for spatial-temporal world modelling. \ac{FACTS}  conceptualises the input as a set of nodes and introduces a permutable memory that can incorporate complex structures. Through selective memory-input routing, input features are dynamically assigned to distinct state-space factors, i.e., explanatory latent representations, that capture the underlying dynamics of the system. This formulation ensures input permutation invariance in the state-space memory, allowing the model to learn consistent factor representations over time, even when the spatial or temporal relationships between input features and factors change as illustrated in Figure~\ref{fig:facts}. Additionally, by treating inputs as a set of nodes, FACTS: i) can incorporate a wider range of input structures, e.g., images or graphs or sets. ii) maintains consistent representations of inputs (e.g., agents), regardless of their order. This allows the model to capture regularities in both spatial/local modalities and temporal dependencies, enhancing its memory efficiency and long-term prediction capabilities through more efficient history compression~\citep{schmidhuber1992learning,schmidhuber2003exploring}. To validate our proposed world modelling approach, we conduct an extensive empirical analysis across multiple tasks, such as multivariate time series forecasting and object-centric world modelling, demonstrating that FACTS consistently matches or exceeds the performance of specialised state-of-the-art models. This confirms its robustness and versatility in addressing complex high-dimensional sequential tasks.

In summary, our main contributions are as follows:
\begin{itemize}[leftmargin=*,topsep=0.1em,itemsep=-0.2em]
 \item  We introduce  \acf{FACTS}, a novel recurrent framework that incorporates a permutable memory structure, enabling flexible and efficient modelling of complex spatial-temporal dependencies.
 \item  FACTS dynamically assigns input features to distinct latent state-space factors, ensuring effective history compression and enhancing long-term prediction power.
\item We formally and empirically show that FACTS achieves consistent factor representations over time, regardless of changes in the spatial order of input features over time, providing robustness in dynamically evolving environments.
\item  We validate the robustness and predictive power of FACTS through extensive world modelling experiments, demonstrating its superior or competitive performance across multivariate time-series forecasting, object-centric world modelling, and spatial-temporal graph prediction tasks.
\end{itemize}

\section{Preliminaries}\label{prelmn}
Structured-state space Models (SSMs)  have their roots in the classic Kalman filter \citep{kalman1960new}, where they process a $m$-dimensional input signal $\vx(t) \in \R^m$ into a $d$-dimensional latent state $\vz(t) \in \R^d$, which is then projected onto an output signal $\vy(t) \in \R^n$. The general form of an SSM is expressed as follows:
\begin{align}
    \dot{\vz}(t) &= \mA(t)\vz(t) + \mB(t)\vx(t) \\
    \vy(t) &= \mC(t)\vz(t) + \mD(t)\vx(t),
\end{align}
where $\dot{\vz}(t) = \frac{d}{dt} \vz(t)$ indicates the time derivative of the state. The matrices $\mA(t) \in \mathbb{R}^{d \times d}$, $\mB(t) \in \mathbb{R}^{d \times m}$, $\mC(t) \in \mathbb{R}^{n \times d}$, and $\mD(t) \in \mathbb{R}^{n \times m}$ present the state, input, output, and feed-forward matrices, respectively. In systems without direct feedthrough, $\mD(t)$ becomes a zero matrix.  Furthermore, since the original system operates in a continuous domain, discretisation is often used~\citep{wang2024state,smith2022simplified}, resulting in the general discrete-time formulation of SSM: 
\begin{align} 
    \vz_t &= \overline{\mA}_t \vz_{t-1} + \overline{\mB}_t \vx_t \label{discrit_eqs1}\\
    \vy_t &= \mC_t \vz_t \label{discrit_eqs2}
\end{align}
with $\overline{\mA}_t$, $\overline{\mB}_t$, and $\mC_t$  govern the dynamics driven by the input sequence $\vx_{\leq t}$, with different constructions~\citep{wang2024state,gu2023mamba,dao2024transformers} influencing the expressiveness and efficiency of the model. If we denote the state vector with $\vh$, we can see that \eqref{discrit_eqs1}-\eqref{discrit_eqs2} form is equivalent to the RNN dynamics. Hence, similarly to RNNs, the system, in \eqref{discrit_eqs1}-\ref{discrit_eqs2}, is inherently sequential, which limits parallel processing.


\tbf{Parallelisation and the selective mechanism} As shown in~\cite{blelloch1990prefix,smith2022simplified} if $\Bar{\mB}_t$ constructed independently of $\vz_{t-1}$, the linear recurrence in ~\eqref{discrit_eqs1} can be expanded as~\eqref{eq:dssm_z_paraf}: 
\begin{align}
    \vz_{t} &= \sum^t_{s=0} \Bar{\mA}_{t:s}^{\times} \Bar{\mB}_s \vx_s,
    \label{eq:dssm_z_paraf} 
\end{align}
where $\Bar{\mA}_{t:s}^{\times}:=\Bar{\mA}_{t+1}...\Bar{\mA}_{s+2}\Bar{\mA}_{s+1}$; $\Bar{\mA}_{t+1} = \mI$; $\Bar{\mB}_0 \vx_0 = \vz_0$ with respect to some initialisation. This expansion not only allows for parallel computation of the linear terms, but also reveals the direct connection established between distant inputs/observations along the sequential dimension, e.g. $\vx_0$ and $\vx_t$ with $t \gg 0$, thereby facilitating the capture of long-term dependencies. Furthermore, integrating a selective mechanism~\citep{gu2023mamba} by constructing $\Bar{\mA}_t$ and $\Bar{\mB}_t$ as functions of each input leads to the following formulation:
\begin{align}
    \vz_{t} &= \Bar{\mA}(\vx_t) \vz_{t-1} + \Bar{\mB}(\vx_t) \vx_t \label{eq:dssm_z_self} \\
            &= \sum^t_{s=0} \Bar{\mA}^{\times}(\vx_{t:s}) \Bar{\mB}(\vx_s) \vx_s  
                 \label{eq:dssm_z_para_sel} \\
    \vy_t   &= \mC(\vx_t) \vz_t \label{eq:dssm_y_sel}
\end{align}
This formulation enables content-aware compression of the historical information, addressing the issue of memory decay in long-sequence modelling. Such selective mechanism is foundational to the effectiveness of modern state-space models~\citep{gu2023mamba,dao2024transformers}. Additionally, as indicated in Eq.~\eqref{eq:dssm_z_self}, SSMs can support parallel computation, since the term $\Bar{\mB}(\vx_t) \vx_t$ remains independent of the preceding state $\vz_{t-1}$.

\section{Proposed Framework: FACTS}  \label{facts}
In this section, we introduce our proposed framework: \textbf{FACTS} (\tbf{FACT}ored \tbf{S}tate-space) model, a novel class of recurrent neural networks designed with a structured state-space memory. FACTS is characterised by two key features: \textbf{permutable state-space memory}, which allows for flexible representation of system dynamics with more complex structures and \textbf{invariant recurrence} with respect to permutations of the input features, ensuring consistent modelling of underlying factors in the world across different time steps.

One key intuition behind the ``permutable state-space memory'' in FACTS is the principle of \textit{history compression} \citep{schmidhuber1992learning,schmidhuber2003exploring},  which emphasises the need to eliminate redundant information in sequence modelling while uncovering algorithmic regularities. This principle is essential for effective long-sequence modelling with high-dimensional data, as it improves generalisation by reducing the accumulation of unnecessary information. Existing SSMs often address this challenge by imposing fixed structural constraints on their state spaces, such as diagonal or block-diagonal structures, to capture invariant components that persist throughout the sequence~\citep{gu2023mamba,gupta2022diagonal,gupta2022simplifying,dao2024transformers}.  However, these fixed structural priors assume that specific dimensions of the state space correspond to consistent and specific factors over time. This assumption can be limiting in world-modelling scenarios where the relationship between state-space dimensions and input features evolves dynamically. For instance, in video sequence modelling, factors may correspond to moving objects, and the spatial location of these objects (i.e., pixel positions) changes from frame to frame. In such cases, the model needs to adapt to these changes, but current SSMs formulations struggle to maintain consistent factor representations due to their rigid structural constraints, i.e., in \eqref{eq:dssm_z_para_sel} the matrices $\Bar{\mB}(\vx_t)$ must not only select relevant information for modelling sequence dynamics but also account for the changing relative orders between subspaces of $\vz_{t-1}$ and $\vx_t$, which can evolve over time.  This introduces additional complexity, leading to noise and redundancy that can hinder effective history compression.   

\subsection{FACTS formulation} 
\vspace{-0.5em}
The FACTS framework is formalised as a class of structured state-space model, which can capture the dynamic interactions between the latent factors and the input features.  To facilitate this dynamic \textit{factorisation}, i.e.,  the process of identifying and disentangling meaningful factors from the input data over time,  at each time step $t$,
we conceptualise the hidden state $Z_t$ as a graph and hence, the state-space memory is represented as a set of nodes that correspond to the latent factors. The input features $X_t$ are also treated as another set of nodes. Formally, let
\begin{align}
Z_t = \{ \vz^1_t, \vz^2_t, \dots, \vz^k_t \} \hspace{2 cm } X_t = \{ \vx^1_t, \vx^2_t, \dots, \vx^m_t \}
\end{align}
where $Z_t$ denote the set of $k$ latent factors at time step $t$ and $m$ is the number of input features. By formulating both sets as nodes, FACTS is inherently invariant to permutations of both input features and factors. Then, to efficiently learn the optimal connections between these two sets, we propose a graph-based routing mechanism that can  effectively match input features with the corresponding factors .i.e., learn edges between nodes in $Z_t$ and $X_t$, reflecting the strength of each correspondence between each input feature and each latent factor.

\textbf{Dynamic selective state-space updates} In analogy with the standard SSM dynamics ~\eqref{discrit_eqs1}-\eqref{discrit_eqs2},  the evolution of the latent factors of FACTS is governed by the following modified state-space dynamics:
\begin{align}
    Z_t  &= \Bar{\mA}_t \odot Z_{t-1} + \Bar{\mB}_t \odot \mU_t  \label{eq:facts_ssm}  \\
   \vy_t &= Dec(\mC_t \odot Z_t)  \label{eq:facts_y}  
\end{align}
Here, $ Z_t$ represents the state-space memory at time $ t $, which stores the latent factors. The terms $ \Bar{\mA}_t $, $ \Bar{\mB}_t $, and $\mC_t$ are selective state-space model parameters responsible for controlling the information flow between the previous memory $Z_{t-1}$ and the input features $X_t$. The symbol $ \odot $ denotes element-wise multiplication, while $Dec$ is a permutation-invariant decoder applied to the latent factors.

Compared to the standard SSM dynamics, i.e.,~\eqref{discrit_eqs1}-\eqref{discrit_eqs2}, we note two key differences: (i) FACTS relies on element-wise multiplication, instead of matrix multiplication, to conserve the invariance properties. (ii) $x_t$ in ~\eqref{discrit_eqs1} is replaced with $\mU_t=(Z_{t-1}, X_t)$, which is a key element in FACTS that models the interactions between the memory $Z_{t-1}$ and the input features $X_t$.

\tbf{The attention-based router} Before diving into the details of the different parts of \eqref{eq:facts_ssm} and \eqref{eq:facts_y}, we first introduce the routing mechanism used in this work. To maintain the recurrent permutability of~\eqref{eq:facts_ssm}, the routing mechanism between memory and inputs must dynamically assign input features to consistent factors. This can be done using an attention-based routing mechanism defined as follows:
\begin{align}
    Z_{t-1} \lcirclearrowright_{\phi, \psi, \varphi} X_t = \softmax\left(\frac{\phi(Z_{t-1})\psi^T(X_t)}{\sqrt{d}}\right) \varphi(X_t)  \label{eq:attn_rout}
\end{align}
where the operator $ \lcirclearrowright $ learns the relationships between the memory $ Z_{t-1} $ and input features $ X_t $, dynamically determining which features correspond to which latent factors. The functions $\phi$, $\psi$, and $\varphi$ represent the query, key, and value mappings, respectively, and are applied row-wise to the memory and input features. 


\textbf{Factorisation process} The term $ \mU_t = \mU(Z_{t-1}, X_t)$ in \eqref{eq:facts_ssm} is crucial for capturing the interactions between the memory and the input features. Note that in prior works~\citep{gu2023mamba,dao2024transformers} $ \mU_t$ is typically constructed as function of the current input $X_t$ only. In this paper, we argue that, similar to the gating in RNNs vs LSTMs~\citep{hochreiter1997long}, it is more effective use both $X_t$ and $Z_{t-1}$ to conserve long term dependencies. This interaction plays a key role in factorisation, which refers to the process of binding the input features to specific memory items, effectively uncovering the underlying factors. In the FACTS framework, the memory at the previous time step $ Z_{t-1}$ serves as the prior over the latent factors, and $\mU_t$ is the factor momentum that guides the evolution of these factors across time. This factor momentum is computed as:
\begin{align}
    \mU_t = Z_{t-1} \lcirclearrowright_{\phi_U, \psi_U, \varphi_U} X_t
\end{align}
where $\phi_U$, $\psi_U $, and $\varphi_U $ are its corresponding query, key, and value mappings.

\tbf{Selectivity through memory-input routing} The selective state-space model parameters $ \Bar{\mA}_t $, $ \Bar{\mB}_t $, and $ \mC_t $ are constructed through interactions between the memory and the input features, ensuring that both the memory and the inputs jointly decide which information should be retained or updated. These parameters are computed as follows:
\begin{align}
 \Delta_t &= Z_{t-1} \lcirclearrowright_{\phi_{\Delta}, \psi_{\Delta}, \varphi_{\Delta}} X_t \hspace{3.1cm}     \Bar{\mA}_t = \exp(\alpha \Delta_t) \\
    \Bar{\mB}_t &= \Delta_t \odot (Z_{t-1} \lcirclearrowright_{\phi_{B}, \psi_{B}, \varphi_{B}} X_t) \hspace{2cm}
    \mC_t = Z_{t-1} \lcirclearrowright_{\phi_{C}, \psi_{C}, \varphi_{C}} X_t
\end{align}
Here, $\Delta_t$ is a step size introduced for discretisation, and the functions $ \phi_{\Delta} $, $\psi_{\Delta} $, and $ \varphi_{\Delta} $ (as well as their counterparts for $ \Bar{\mB}_t $ and $ \mC_t $) are responsible for mapping the memory and inputs to their respective selective parameters. The exponential function $ \exp $ ensures that the selective parameters are non-negative, while $ \alpha $ is a trainable scalar controlling the influence of $ \Delta_t $. By employing this selective mechanism, FACTS is capable of compressing long sequences in its state-space memory while maintaining the key properties of latent permutation equivariance and row-wise permutation invariance. Hence, FACTS can efficiently capture meaningful factors, e.g., objects in video frames or independent sources in a signal,  even as their relationships with input features change over time. 

\textbf{Linearisation} Although the framework presented so far in \eqref{eq:facts_ssm}  has a permutable state-space memory and equivariant through the memory-input routing, which we formally show in Section~\ref{sec:facts_theory}, the routing between $Z_{t-1}$ and $X_t$ introduces dependencies of $\Bar{\mB}(Z_{t-1}, X_t)$ and $\mU(Z_{t-1}, X_t)$ on $Z_{t-1}$. This results in a non-linear recurrence between $Z_{t-1}$ and $Z_t$ in~\eqref{eq:facts_ssm}, which limits parallelisation and hinders training efficiency. To overcome this, we substitute $Z_{t-1}$ with $Z_0$, i.e. the initial memory or state, within the information routing processes. This leads to the final formulation of FACTS:
\begin{align}
    Z_t  &= \FACTS(Z_{t-1}, Z_0, X_t)\\ &= \Bar{\mA}(Z_0, X_t) \odot Z_{t-1} + \Bar{\mB}(Z_0, X_t) \odot \mU(Z_0, X_t)  \label{eq:pfacts_rnn} \\ 
    &= \Bar{\mA}(Z_0, X_t) \odot \FACTS(Z_{t-2}, Z_0, X_{t-1}) + \Bar{\mB}(Z_0, X_t) \odot \mU(Z_0, X_t)\label{eq:pfacts_rr}\\
         &= \sum^t_{s=0} \Bar{\mA}^{\times}(Z_0, X_{t:s})\odot \Bar{\mB}(Z_0, X_s) \odot \mU(Z_0, X_s),   \label{eq:pfacts_parallel}\\
         &= \FACTS(Z_0, X_{1:t})   \label{eq:pfacts_t}  
\end{align}
where $\Bar{\mA}^{\times}(Z_0, X_{t:s})=\Bar{\mA}_{t+1}\odot \Bar{\mA}_{t}\odot\Bar{\mA}_{t-1}...\odot \Bar{\mA}_{s+1}$ and $\Bar{\mA}_{t+1}$ is filled with ``1''; the initial state $Z_0$ can be given a priori, learnable, or sampled from a prior distribution. This formulation in~\eqref{eq:pfacts_rnn} linearise the recurrence in~\eqref{eq:facts_ssm} by breaking the non-linear dependency between $Z_t$ and $Z_{t-1}$. That is, as shown in~\eqref{eq:pfacts_t}, the inputs interact only with the initial memory, enabling fast computation of $Z_t$ using~\eqref{eq:pfacts_parallel}, i.e., without recurrence. This significantly improves computational efficiency. Note that while $Z_0$ is denoted as the ``initial state'', it needs not represent the sequence’s true start; long sequences can be segmented (chunked) for parallel computation within segments while maintaining sequential dependencies across the original sequence.

\subsection{Theoretical Analysis of FACTS} \label{sec:facts_theory}
\vspace{-0.5 em}
Here, we formally proof the permutation equivariance and invariance properties of FACTS.  We first formally define the two fundamental properties, namely \textit{left permutation equivariant} (L.P.E.) and  \textit{right permutation invariant }(R.P.I.) in Definitions~\ref{def:LPE} and ~\ref{def:RPI}, respectively.
\begin{definition} \label{def:LPE}
Let $f: \R^{n_1\times n_2} \times \R^{t \times n_3 \times n_4} \to \R^{n_1 \times n_5}$ be a bivariate function with $n_1, n_2, n_3, n_4, n_5, t \in \mathbb{N}$. $f$ is permutation equivariant (L.P.E.)  if for all $\sigma \in S_{n_1}, \mM_1\in \R^{n_1 \times n_2},$ and $\mM_2 \in \R^{t\times n_3 \times n_4}$,
\begin{align*}
    f(\sigma \mM_1, \mM_2) = \sigma f(\mM_1, \mM_2),
\end{align*}
where $S_k$ denotes the set of permutation matrices of size $\R^{k \times k}$.
\end{definition}
\begin{definition}  \label{def:RPI}
    Let $\R^{n_1\times n_2} \times \R^{t \times n_3 \times n_4} \to \R^{n_1 \times n_5}$ be a bivariate function with $n_1, n_2, n_3, n_4, n_5, t \in \mathbb{N}$. $f$ is right permutation invariant (R.P.I.) if for all $\sigma_1, \sigma_2, \dots, \sigma_t \in S_{n_3}, \mM_1\in \R^{n_1\times n_2},$ and $\mM_2^1, \mM_2^2, \dots, \mM_2^t \in \R^{n_3\times n_4}$,
\begin{align*}
    f(\mM_1, [\sigma_1 \mM_2^1, \sigma_2 \mM_2^2, \dots, \sigma_t\mM_2^t]) = f(\mM_1, [ \mM_2^1,  \mM_2^2, \dots, \mM_2^t]).
\end{align*}
\end{definition}
These \LPE~and \RPI~properties, which formally describe the two fundamental aspects of FACTS: \textit{permutable memory} and \textit{permutation-invariant recurrence (w.r.t. the features)} --- with memory $Z_{t-1}$ and input $X_t$ serving as the left and right arguments of FACTS. They are thus essential not only for constructing the routing mechanism but also for the overall design of FACTS. 

Using Definitions~\ref{def:LPE} and~\ref{def:RPI}, that by taking memory $Z_t$ and features $X_t$ as the left and right arguments in \ac{FACTS} (\eqref{eq:pfacts_t}), we can show the following result:
\begin{theorem}
\label{therm:facts}  
$\FACTS$ as defined in~\eqref{eq:pfacts_t} is L.P.E. and R.P.I. 
\end{theorem}
The proof of Theorem~\ref{therm:facts} is available in Appendix~\ref{append:proofs}. Theorem~\ref{therm:facts}  proves our main claim that FACTS: i) is invariant to input features permutation. ii) learns permutable state-space memory.  Furthermore, it is possible to extend our results in Theorem~\ref{therm:facts} to the more general case, where $\Bar \mA, \Bar \mB, \mU$ are L.P.E. and R.P.I. functions of $Z_{t-1}$ and $X_t$. The main result is presented in Theorem~\ref{therm:generalised}.
\begin{theorem}
\label{therm:generalised}  
if $\Bar \mA, \Bar \mB, \mU$ are L.P.E. and R.P.I. functions of $Z_{t-1}$ and $X_t$, any dynamics governed by~\eqref{eq:facts_ssm} is L.P.E. and R.P.I. 
\end{theorem}
The proof of Theorem~\ref{therm:generalised} is available in Appendix~\ref{append:proofs}. Theorem~\ref{therm:generalised} highlights the main condition on the variables $\Bar \mA, \Bar \mB, \mU$ to ensure that the model is invariant to input features and has an equivariant memory. This can spark future research to develop SSM models based on~\eqref{eq:facts_ssm} that are efficient history compressors and are suitable to dynamic world modelling scenarios.

\section{Experiments}\label{exp}
We design experiments to evaluate the effectiveness of FACTS in world modelling. We frame world modelling as a prediction task, where the model must predict future events in complex environments based on observed history, and we assess its performance by prediction accuracy. Experiments are conducted on three environments: the multivariate time series (MTS) benchmark for forecasting, synthetic multi-object videos~\citep{yiclevrer,greff2021kubric,lin2020gswm}, and dynamic-graph node prediction~\citep{li2018diffusion}. Additionally, Appendix~\ref{ablation_app} presents an extensive ablation study that highlights the robustness of FACTS.


\subsection{Long term multivariate time-series forecasting}
\vspace{-0.5em}
In many real-world applications, such as climate prediction, traffic flow management, or autonomous systems, predicting future states over long horizons is crucial for effective decision-making. World modelling, in these domains, often involves high-dimensional multivariate inputs—such as interacting agents, variables, or environmental factors—requiring the system to account for their complex dependencies and interactions across time. Long-term forecasting in this context is challenging as it demands accurate representation of temporal dynamics over extended periods. Additionally, input features may lack a predefined order, or this order could change dynamically. For example, a system may receive data from sensors (e.g., temperature and pressure) without knowing which is which during testing. The world model must provide reliable long-term predictions even if the input order changes unexpectedly and generalise to unseen configurations, without requiring exposure to every possible permutation of input features during training.

We use the open-source Time Series Library (TSLib)\footnote{Time Series Library benchmark: \url{https://github.com/thuml/Time-Series-Library.git}} to evaluate long-term multivariate time-series forecasting (MSTF) tasks across 9 real-world datasets spanning multiple domains (e.g., energy, weather, and finance). FACTS is compared against 8 state-of-the-art baseline models~\citep{wang2024smamba, liu2023itransformer, wu2022timesnet,Yuqietal2023patchtst,zeng2023dlinear,zhang2023crossformer,zhou2022fedformer,wu2021autoformer}, following TSLib’s standardised settings: the input sequence length is fixed at 96, with prediction lengths of \{96, 192, 336, 720\}. Performance is evaluated using mean-squared error (MSE) and mean-absolute error (MAE). C.f. Appendix~\ref{append:LTF} for more details.



\vspace{-0.5em}
\subsubsection{Forecasting with predefined order (Scenario I)} 
\vspace{-0.5em}
We use the exact same setup to~\cite{wang2024smamba, liu2023itransformer,wu2022timesnet}, with the exception of the pre- and post-processing modules (referred to as the ``embedders'' and ``projectors'' in TSLib). In our implementation, we replace these with set functions to accommodate the output structure of \ac{FACTS} (c.f. Appendix~\ref{append:impl} for more details). Note that in the standard setup of TSLib, the arrangement of the input features in the test is not changed and is the identical to the arrangement to the one seen during the training.  The average results over the different prediction windows of our proposed approach along with all competing methods are presented in Table~\ref{table_MSTF} and the full results are available in Table~\ref{tab:full_MTS} in Appendix~\ref{append:add_results}.

\begin{table*}[t]
  \centering
  \setlength{\extrarowheight}{1.pt}
   \resizebox{\textwidth}{!}{\begin{tabular}{l|c|ccccccccc}
   \toprule
		  & & ETTm1 & ETTm2 &ETTh1 &ETTh2& Electricity &Exchange & Traffic & Weather &  Solar-Energy \\ 
     \cmidrule(r){2-11}
\multirow{2}{*}{\textbf{Autoformer} (\citeyear{wu2021autoformer}) }  
    & \textbf{MSE} & 0.588 & 0.327 & 0.496 & 0.450 & 0.227 & 0.613 & 0.628 & 0.338 & 0.885 \\ 
    & \textbf{MAE} & 0.617 & 0.371 & 0.487 & 0.459 & 0.338 & 0.539 & 0.379 & 0.382 & 0.711 \\ 
    \hline
\multirow{2}{*}{\textbf{FEDformer} (\citeyear{zhou2022fedformer}) } 
    & \textbf{MSE} & 0.448 & 0.305 & \textcolor{red}{\textbf{0.440}} & 0.437 & 0.214 & 0.519 & 0.610 & 0.309 & 0.291 \\ 
    & \textbf{MAE} & 0.452 & 0.349 & 0.460 & 0.449 & 0.327 & 0.429 & 0.376 & 0.360 & 0.381 \\ 
    \hline
\multirow{2}{*}{\textbf{TimesNet} (\citeyear{wu2022timesnet}) } 
    & \textbf{MSE} & 0.400 & 0.291 & 0.458 & 0.414 & 0.192 & 0.416 & 0.620 & 0.259 & 0.301 \\ 
    & \textbf{MAE} & 0.406 & 0.333 & 0.450 & 0.427 & 0.295 & 0.443 & 0.336 & 0.287 & 0.319 \\ 
    \hline
\multirow{2}{*}{\textbf{PatchTST} (\citeyear{Yuqietal2023patchtst})}
    & \textbf{MSE} & \textcolor{red}{\textbf{0.387}} & \textcolor{red}{\textbf{0.281}} & 0.469 & 0.387 & 0.205 & 0.367 & 0.481 & 0.259 & 0.270 \\ 
    & \textbf{MAE} & \textcolor{blue}{0.400} & \textcolor{blue}{0.326} & 0.454 & 0.407 & 0.290 & 0.404 & 0.304 & 0.281 & 0.307 \\ 
    \hline
\multirow{2}{*}{\textbf{DLinear} (\citeyear{zeng2023dlinear})} 
    & \textbf{MSE} & 0.403 & 0.350 & 0.456 & 0.559 & 0.212 & \textcolor{blue}{0.354} & 0.625 & 0.265 & 0.330 \\ 
    & \textbf{MAE} & 0.407 & 0.401 & 0.452 & 0.515 & 0.300 & 0.414 & 0.383 & 0.317 & 0.401 \\ 
    \hline
\multirow{2}{*}{\textbf{Crossformer} (\citeyear{zhang2023crossformer})} 
    & \textbf{MSE} & 0.513 & 0.757 & 0.529 & 0.942 & 0.244 & 0.940 & 0.550 & 0.259 & 0.641 \\ 
    & \textbf{MAE} & 0.496 & 0.610 & 0.522 & 0.684 & 0.334 & 0.707 & 0.304 & 0.315 & 0.639 \\ 
    \hline
    \multirow{2}{*}{\textbf{iTransformer}  (\citeyear{liu2023itransformer}) } 
    & \textbf{MSE} & 0.407 & 0.288 & 0.454 & 0.383 & 0.178 & 0.360 & \textcolor{blue}{0.428} & 0.258 & \textcolor{red}{\textbf{0.233}} \\ 
    & \textbf{MAE} & 0.410 & 0.332 & \textcolor{blue}{0.447} & 0.407 & 0.270 & \textcolor{blue}{0.403} & \textcolor{blue}{0.282} & \textcolor{blue}{0.278} & \textcolor{red}{\textbf{0.262}} \\ 
    \hline
    \multirow{2}{*}{\textbf{S-Mamba} (\citeyear{wang2024smamba})} 
    & \textbf{MSE} & 0.398 & 0.288 & 0.455 & \textcolor{blue}{0.381} & \textcolor{blue}{0.170} & 0.367 & \textcolor{red}{\textbf{0.414}} & \textcolor{red}{\textbf{0.251}} & \textcolor{blue}{0.240} \\ 
    & \textbf{MAE} & 0.405 & 0.332 & 0.450 &  \textcolor{blue}{0.405}   & \textcolor{blue}{0.265} & 0.408 & \textcolor{red}{\textbf{0.276}} & \textcolor{red}{\textbf{0.276}} & {0.273} \\ 
    \hline
\multirow{2}{*}{\textbf{FACTS (Ours)}}  
    & \textbf{MSE} & \textcolor{blue}{0.392} & \textcolor{red}{\textbf{0.281}} & \textcolor{red}{\textbf{0.440}} & \textcolor{red}{\textbf{0.373}} & \textcolor{red}{\textbf{0.166}} & \textcolor{red}{\textbf{0.342}} & 0.472 & \textcolor{red}{\textbf{0.251}} & 0.253 \\  
    & \textbf{MAE} & \textcolor{red}{\textbf{0.397}} & \textcolor{red}{\textbf{0.325}} & \textcolor{red}{\textbf{0.428}} & \textcolor{red}{\textbf{0.399}} & \textcolor{red}{\textbf{0.263}} & \textcolor{red}{\textbf{0.392}} & 0.303 & \textcolor{blue}{0.278} & \textcolor{blue}{0.272} \\ 
    \bottomrule
  \end{tabular} }
  \caption{Average MSE and MAE of the different approaches on the multivariate time series forecasting tasks. For each metric and each dataset, the top performance and the second best are highlighted in \textcolor{red}{\textbf{red}} and \textcolor{blue}{blue}, respectively.}
  \label{table_MSTF}
  \vspace{-0.5em}
\end{table*}

\textit{\textbf{Results}}~Table \ref{table_MSTF} highlights the strong performance of FACTS, which achieves competitive results in both metrics, compared to the competing state-of-the-art specialised MSTF models. As shown, FACTS achieves the best MAE and MSE on 6 out of 9 datasets and, in terms of MAE, is always in the top 2 in 8 out of 9 them. Even where it is not the top performer, FACTS remains highly competitive (2nd or 3rd place) as seen in Traffic and Solar-Energy underscoring its robustness and ability to adapt to different scenarios. FACTS' ability to efficiently capture long-term dependencies can be attributed to its structured state-space memory, which promotes the learning of modular patterns whose interactions can explain the spatial-temporal correlations of the multivariate observations. 

\textbf{Parallel vs Recurrent FACTS} The parallelisation design outlined in our~\eqref{eq:pfacts_rnn}-\ref{eq:pfacts_t} not only enhances computational efficiency but also provides flexibility for recurrent applications of the FACTS core operations. As mentioned, long sequences can be segmented for parallel computation within segments while preserving sequential dependencies across the entire sequence. To clarify further, we analysed FACTS’ long-term forecasting performance by comparing its fully sequential mode to its fully parallel mode, adjusted by segment window size, on the MTS Electricity dataset (c.f. Figure~\ref{fig:rnn-vs-parallel}). Each window size represents a different update frequency for $Z_0$: for example, a window size of 1 corresponds to fully recurrent FACTS, while a window size of 96 (the sequence length) corresponds to fully parallel FACTS. As shown in Figure~\ref{fig:rnn-vs-parallel}, the models consistently maintained strong performance across various segment window sizes.

\vspace{-0.5em}
\subsubsection{Forecasting with unknown order (Scenario II)}
\vspace{-0.5em}
To evaluate robustness under dynamic scenarios, we randomly permute the input features during the testing to simulate environments where the arrangement of agents or entities (e.g., robots, sensors) changes unpredictably. This mirrors real-world scenarios where input configurations vary, challenging world models to adapt to unseen input orderings. We focus on top pretrained models from the first scenario (Table ~\ref{table_MSTF}), i.e., FACTS, iTransformer, and S-Mamba. For the datasets, we use the challenging ones from the first scenario and permute the feature embeddings five times during testing, reporting the average performance and two standard deviations.

\begin{figure*}[t]
\centering
\includegraphics[width=0.24\linewidth]{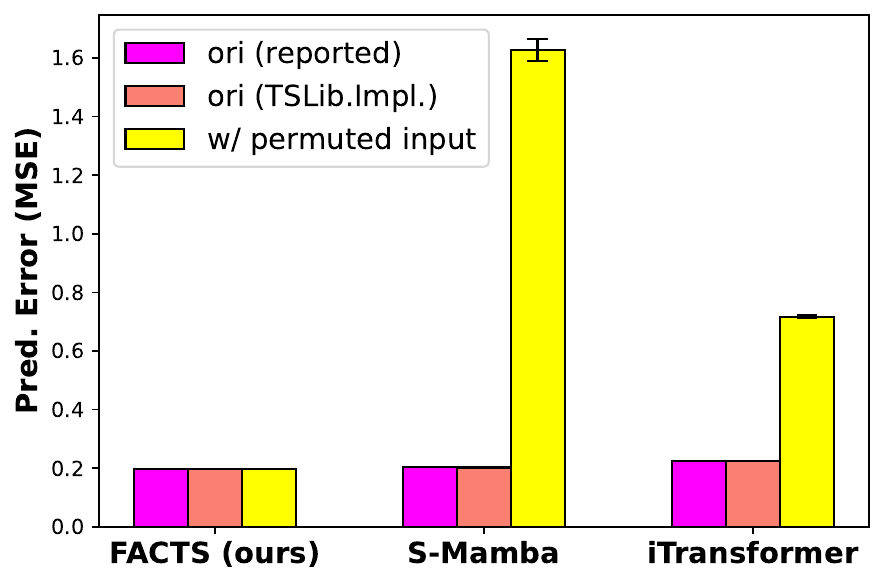}
\includegraphics[width=0.24\linewidth]{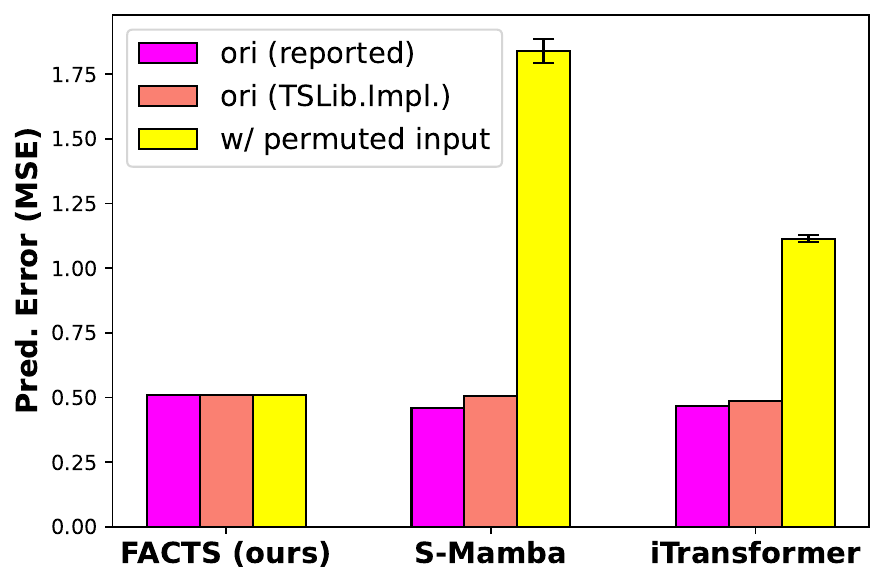}
\includegraphics[width=0.24\linewidth]{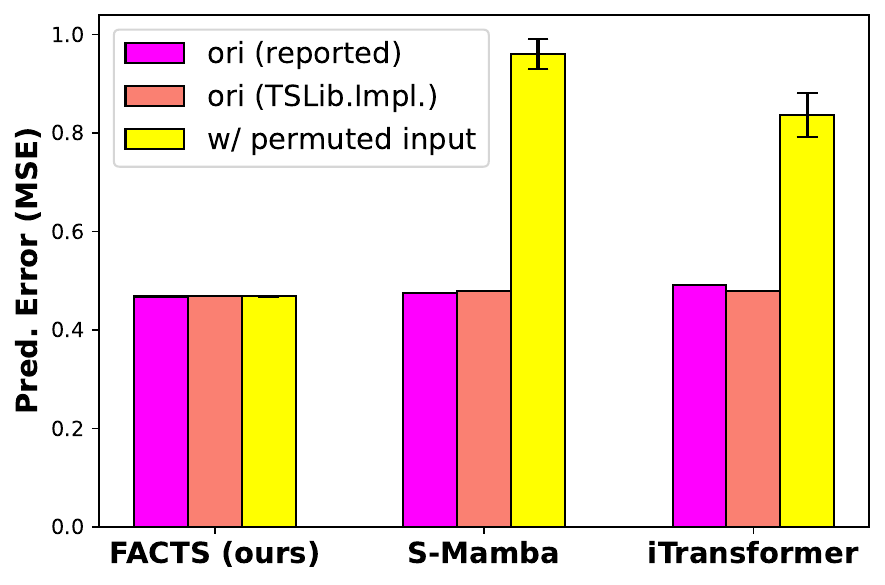}
\includegraphics[width=0.24\linewidth]{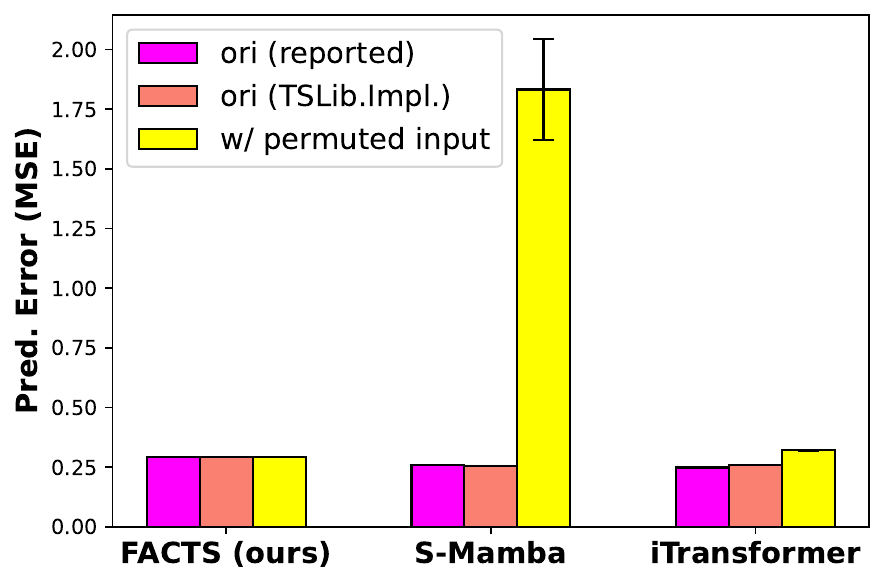}
\caption{Model robustness to input permutations on 4 MTSF datasets (left to right: Electricity, Traffic, ETTm1, SolarEnergy). Magenta bars represent original performance, salmon bars show performance using our/TSLib implementation, and yellow bars represent results under input permutation. Results are averaged over five random seeds, with error bars showing $\pm 2\times$ standard deviation.}
\vspace{-0.5em}
\label{fig:perm_input}
\end{figure*}

\textit{\textbf{Results}}~The main results are presented in Figure~\ref{fig:perm_input}. While other models, iTransformer and S-Mamba, experience significant degradation in performance when the input feature orders are shuffled, FACTS consistently maintains its prediction performance across the different tasks. These findings corroborate the theoretical results of Section~\ref{sec:facts_theory}. For example, on the Traffic dataset, the MSE error of S-mamba, which is the model with the top performance in the standard setting (Table \ref{table_MSTF}), increases more than threefold, and the error for iTransformer doubles. In contrast, FACTS, leveraging its selective memory routing which consistently assigns input features to the latent factors,  preserves low error rates despite the permutation. This highlights FACTS' ability to handle dynamic and unordered environments. This adaptability further emphasizes the generalisation strength of FACTS, particularly in world modelling scenarios where input orders may be inconsistent or unknown. 

\vspace{-0.5em}

\begin{table}[ht]
    \centering
    \begin{minipage}{0.48\textwidth}
        \centering
        \begin{tabular}{l|cc}
            \toprule
                 &  \multicolumn{2}{c}{ LPIPS $\downarrow$ } \\
              
            \textbf{Method} & CLEVRER & OBJ3D  \\
            \midrule
            PredRNN~(\citeyear{wang2017predrnn})
                & 0.17 & 0.12 \\
            VQFormer~(\citeyear{razavi2019vqvae2})        
                & 0.18 &  0.11 \\
            G-SWM~(\citeyear{lin2020gswm})           
                & 0.16 &  0.10 \\
            SlotFormer~(\citeyear{wu2022slotformer})      
                & 0.11 &  0.08 \\
            SAVi-dyn~(\citeyear{kipf2023savi})        
                & 0.19 & 0.12 \\
                \midrule
            \textbf{FACTS (Ours)}   
                & \textbf{0.09} & \textbf{0.07} \\
            \bottomrule
        \end{tabular}
        \caption{Quantitative results for slot dynamics prediction: visual quality of future frame prediction, measured by LPIPS (lower is better).}
        \label{table:slot-dyn-visual}
    \end{minipage}
    \vspace{-1em}
    \hfill
    \begin{minipage}{0.45\textwidth}
        \centering
        \begin{tabular}{l|cc}
            \toprule
                           & FG-ARI$\uparrow$   & mIoU$\uparrow$  \\ 
            \midrule
            SAVi~\citeyear{kipf2023savi}          
                           & 0.64    & 0.43 \\
            OC-SlotSSM~\citeyear{jiang2024slot}    
                           & 0.68    & 0.55 \\
            \midrule
            \textbf{FACTS (Ours)}  
                           & \textbf{0.75} & \textbf{0.60}  \\
            \bottomrule
        \end{tabular}
        \caption{Quantitative results for unsupervised object discovery: segmentation quality on MOVI-A under the video reconstruction setting, evaluated by FG-ARI and mIoU (higher indicate better).}
        \label{table:obj-discovery-movia}
    \end{minipage}
    \vspace{-1em}
\end{table}

\subsection{Object-centric world modelling} 
\vspace{-0.5em}
Visual object-centric representation learning (OCRL)~\citep{burgess2019monet,greff2019multi,nanbo2020learning} tackles the challenge of binding visual information to consistent factors, even as object features dynamically permute with movement across pixels in videos~\citep{kipf2023savi}. This aligns with FACTS’ objective of identifying regularities in dynamic environments for history compression and future-event prediction, making OCRL an ideal evaluation benchmark. To evaluate how FACTS 1) leverages object information for future predictions and 2) aligns its discovered factors with objects, we conduct two OCRL experiments, \textit{slot dynamics prediction} and \textit{unsupervised object discovery}, set on widely-used OCRL datasets~\citep{yiclevrer, lin2020gswm, greff2021kubric}.



\textbf{Slot Dynamics Prediction} 
The task involves having a world model capture object-centric dynamics in latent space: given the latent object representations of observed events (``burn-in''), the model predicts the future latent codes of the objects (``roll-out''). We conducted this experiment following the setup of our major baseline, SlotFormer~\citep{wu2022slotformer}. We evaluate the performance of the model by assessing 1) the visual quality of the predicted future frames and 2) the precision of future segmentation map rendered from the predicted latents. We quantify visual quality using the LPIPS metric, which provides stronger alignment with human perception than other commonly used metics such as PSNR and SSIM~\citep{wu2022slotformer,sara2019image}, and segmentation accuracy using the commonly used Mean Intersection over Union (mIoU) and Adjusted Rand Index (ARI), with and/or without the foreground focus.  

\textit{\textbf{Results}}~The results presented in Table~\ref{table:slot-dyn-visual} and Table~\ref{table:slot-dyn-seg}  highlight the strengths of the FACTS model in terms of both visual quality and segmentation accuracy for object dynamics prediction on the CLEVRER dataset. FACTS achieves the lowest LPIPS score of 0.09, indicating superior visual quality in the predicted frames. Additionally, it achieves competitive segmentation accuracy,  achieving a leading FG-mIoU of $48.11\%$. This highlights its effectiveness in predicting object positions and interactions in future frames. We attribute these results to FACTS’ selective history compression mechanism. In contrast to SlotFormer, which predicts the next state by attending to all past inputs---resulting in inefficiency and noisy predictions---FACTS effectively compresses and retains only the most relevant information in memory, thereby filtering out noise and enabling more accurate dynamics modelling and future predictions.


\textbf{Unsupervised Object Discovery} In contrast to the slot dynamics task, where object slots or factors are given as input, this experiment requires FACTS to automatically discover relevant factors for future predictions in multi-object videos. This process enables us to understand the regularities that FACTS conceptualises as significant for forecasting future events. We utilised a CNN encoder to convert the input image into a feature set, from which FACTS learns the object factors. These factors are employed to predict future object slots and are subsequently decoded back into video frames using a spatial-broadcast decoder. Note that, in this experiment, we adopted the fully-unsupervised setup of SAVi and jointly train all the modules (including the CNN encoder and spatial-broadcast decoder) end-to-end from scratch by minimising the reconstruction MSE and future prediction MSE. In addition to future prediction, we also conducted object discovery under the video reconstruction setting using the MOVi-A dataset to enable direct comparison with the state-of-the-art video-based OCRL models for videos, i.e., SAVi~\citep{kipf2023savi} and OC-SlotSSM~\citep{jiang2024slot}. 

\textit{\textbf{Results}}~We visualise the discovered factors by independently rendering each factor’s dynamics back into videos. In the ``discovery for prediction'' task, FACTS primarily identifies moving objects - considered ``useful'' for future predictions---while treating static objects as background. In contrast, for the ``discovery for reconstruction'' task, FACTS identifies also static objects as explanatory factors (see Figure~\ref{fig:facts-unsup})—like SAVi and OC-SlotSSM. We attribute this behaviour to the residual design of the FACTS predictor, which is muted during reconstruction. Our quantitative results in Table~\ref{table:obj-discovery-movia} demonstrate that FACTS outperforms both OC-SlotSSM and SAVi in unsupervised object discovery (for reconstruction), highlighting its superior performance despite being a more general framework. Additional visual results of our object-centric world modelling are available in the Appendix~\ref{append:add_results}. 


\begin{figure}[t]
\centering
\begin{minipage}[h]{0.52\textwidth} 
    \centering
    \begin{tabular}{lccr}
        \toprule
        \textbf{Method} & \textbf{RMSE}$\downarrow$ & \textbf{MAE}$\downarrow$  &  \textbf{MAPE}$\downarrow$ \\
        \midrule
        HA~\citeyear{li2018diffusion}              & 12.16        & 5.92        & 15.17\%       \\
        LSTNet~\citeyear{lai2018modeling}         & 9.22         & 5.11        & 12.56\%       \\
        STGCN~\citeyear{ijcai2018p505}           & 7.92         & 3.87        & 10.05\%       \\
        DCRNN ~\citeyear{li2018diffusion}      & 7.87         & 3.85        & 10.01\%       \\
        GWN~\citeyear{ijcai2019p264}             & 7.66         & 3.54        & 9.98\%        \\
        ASTGCN~\citeyear{guo2019attention}          & 7.99         & 3.94        & 10.12\%       \\
        GMA~\citeyear{zheng2020gman} & 8.32 & 4.06& 10.91\% \\ 
        MTGNN~\citeyear{wu2020connecting}           & 8.16         & 3.99        & 10.28\%       \\
        AGCRN~\citeyear{bai2020adaptive}           & 8.22         & 4.02        & 10.53\%       \\
        DGCRN~\citeyear{li2023dynamic}           & 7.19          & 3.44         & 9.73\%        \\
        STGM~\citeyear{lablack2023spatio}            & 7.10         & 3.23        & 9.39\%        \\
        MegaCRN~\citeyear{jiang2023spatio} & 7.23 & 3.38 & 9.72\% \\ 
        TESTAM~\citeyear{lee2024testam} & 7.09 & 3.36 &  9.67\% \\
         \midrule
        \textbf{FACTS (Ours)}   & \textbf{6.97} &  \textbf{3.11}  & \textbf{9.08\%} \\ 
    \bottomrule
    \end{tabular}
    \captionof{table}{RMSE, MAE, and MAPE of various approaches for long-term graph node forecasting (1-hour, 12-step ahead) on the METR-LA dataset.}
    \label{table:graph-results}
\end{minipage}%
\vspace{-0.5em}
\hfill 
\begin{minipage}[h]{0.44\textwidth}
    \centering
    \includegraphics[width=\linewidth]{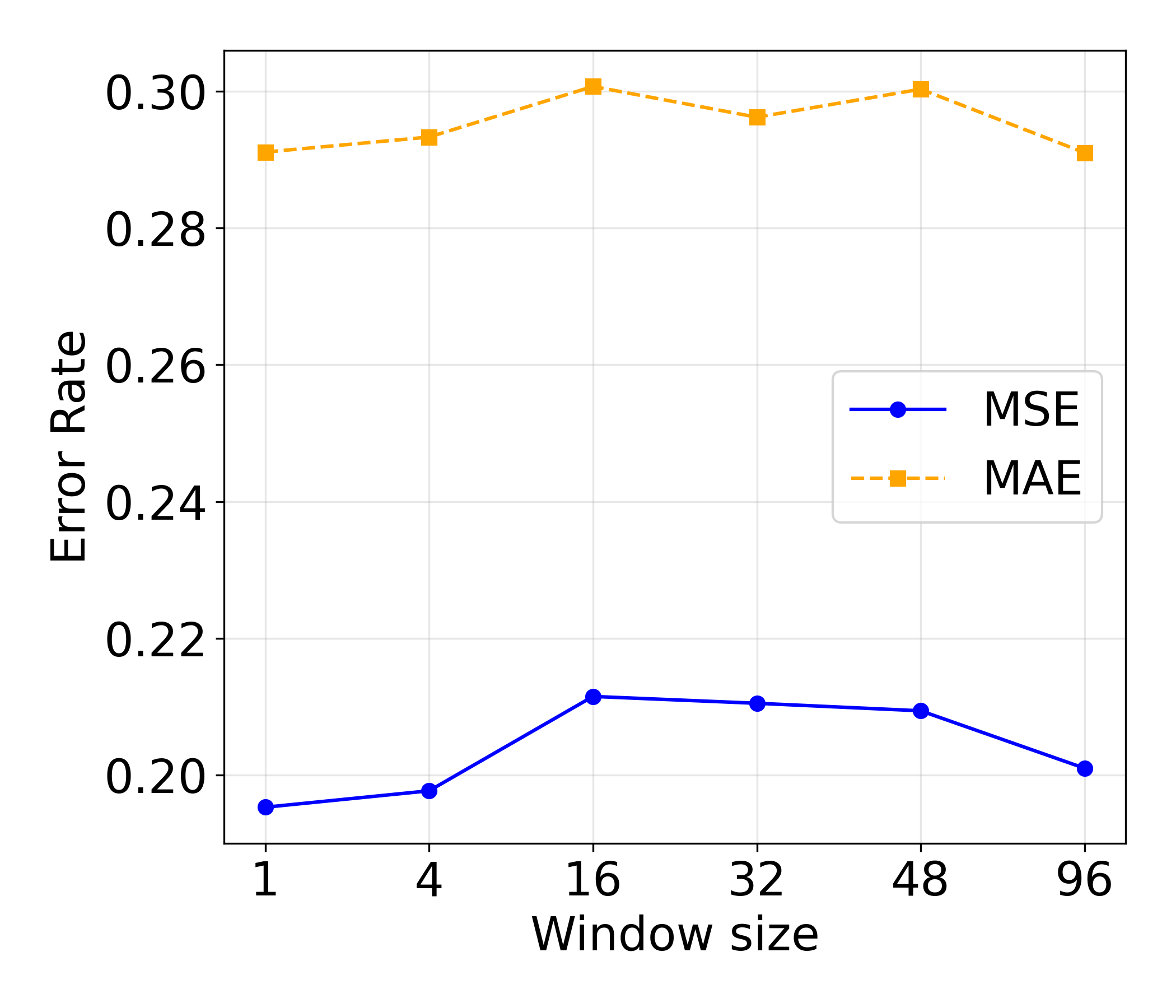} 
    \caption{Parallel vs Recurrent FACTS on long-term MTS (Electricity): MSE and MAE for different window/chunk sizes. Input (observable) sequence length set to 96.}
    \label{fig:rnn-vs-parallel}
\end{minipage}
\vspace{-0.5em}
\end{figure}

\subsection{Long term prediction with graph data} 
\vspace{-0.5em}
To demonstrate FACTS' flexibility as a powerful general framework able to handle different input types/modalities and efficiently solve diverse forecasting-based tasks, we also apply FACTS to dynamic-graph input data evaluated on long-term node prediction task (12-step MAE) with the commonly used METR-LA dataset~\citep{li2018diffusion} and we compare against existing state-of-the-art approaches on this task. We refer to Appendix~\ref{append:graph-pred} for more experimental details.

\textit{\textbf{Results}}~As can be seen in Table~\ref{table:graph-results}. FACTS, leveraging its graph-structured memory, also outperforms all existing methods on this task, even those specialised for this task. For instance, TESTAM~\citep{lee2024testam} yields an MAPE of $9.67\%$ whereas FACTS yields $9.08\%$.
This further corroborates our main claim that FACTS is indeed a versatile world model framework with consistently strong performance in several diverse forecasting tasks.

\section{Discussions \& Conclusion}\label{discuss}
In this work, we introduced FACTS, a novel recurrent framework designed for spatial-temporal world modelling. FACTS is constructed permutable state-space memory, which offers the flexibility needed to capture complex dependencies across time and space. By employing selective memory-input routing, FACTS is able to dynamically assign input features to distinct latent factors, enabling more efficient history compression and long-term prediction accuracy.  Furthermore, we formally showed that FACTS: i) is invariant to input feature permutation. ii) learns permutable state-space memory, maintaining consistent factor representations regardless of changes in the input order. Furthermore, through comprehensive empirical evaluations, FACTS demonstrated superior performance on a variety of real-world datasets, consistently matching or outperforming specialised state-of-the-art models in diverse tasks. Notably, FACTS maintained its predictive powers even in challenging settings where the order of input features was shuffled, highlighting its robustness and adaptability. These results underscore the model’s potential for a wide range of applications, particularly in world modelling scenarios where input configurations are variable or uncertain. For future work, we plan to extend FACTS to larger-scale experiments, exploring its scalability and potential in even more complex world modelling tasks.

\newpage

\section*{Acknowledgement}\label{ack}
The research reported in this publication was supported by funding from King Abdullah University of Science and Technology (KAUST) - Center of Excellence for Generative AI, under award number 5940.

\bibliography{iclr2025_conference}
\bibliographystyle{iclr2025_conference}


\newpage
\appendix

\section{Related Works}\label{append:review}

From a historical perspective, ``world models'' or using models to learn environmental dynamics and leveraging them in policy training has an extensive literature, with early foundations laid in the 1980s using feed-forward neural networks~\citep{werbos1987learning,werbos1989neural,munro1987dual,robinson1989dynamic,nguyen1990truck} and in the 1990s with RNNs~\citep{schmidhuber1990making,schmidhuber1990line,schmidhuber1991possibility,schmidhuber1990reinforcement}. Notably, PILCO~\citep{deisenroth2011pilco,mcallister2016data} has emerged as a key probabilistic model-based method, using Gaussian processes (GPs)~\citep{mackay1998introduction} to learn system dynamics from limited data and train controllers for tasks like pendulum swing-up and unicycle balancing. While GPs perform well with small, low-dimensional datasets, their computational complexity limits scalability in high-dimensional scenarios. To address this, later works~\citep{gal2016improving,depeweg2016learning} have adopted Bayesian neural networks~\citep{kononenko1989bayesian}, which have demonstrated success in control tasks with well-defined states~\citep{hein2017benchmark}. However, these methods remain limited when modelling high-dimensional environments, such as sequences of raw pixel frames. In the context of reinforcement learning, using recurrent models to learn system dynamics from compressed latent spaces has significantly improved data efficiency~\citep{schmeckpeper2020learning,finn2016deep}. While the development of internal models for reasoning about future states using RNNs dates back to the early 1990s, subsequent works, such as ``Learning to Think''~\citep{schmidhuber2015learning} and ``World Models'' \citep{ha2018world}, have extended this by introducing RNN-based frameworks that model environments and reason about future outcomes. These RNN-based models have been applied to future frame generation~\citep{chiappa2017recurrent,oh2015action,denton2017unsupervised} and reasoning about future outcomes~\citep{silver2017predictron,watters2017visual}. However, as RNNs suffer from the
vanishing gradients problem~\citep{hochreiter1991untersuchungen,pascanu2013difficulty}, recently there has been a growing interest in using Transformers~\citep{chen2022transdreamer,robine2023transformer,micheli2022transformers} and SSM-based appraoches~\citep{gu2023mamba,hafner2023mastering,samsami2024mastering} for world modelling.

As world modelling is fundamentally intertwined with sequence modelling~\citep{schmidhuber1990making}, it often carries temporal implications that align with the principle of history compression~\citep{schmidhuber1992learning,schmidhuber2003exploring}. Temporal selectivity is essential in these models, with Recurrent Neural Networks (RNNs), particularly those with gating mechanisms like LSTMs~\citep{hochreiter1997long}, GRUs~\citep{cho2014gru}, and xLSTMs~\citep{beck2024xlstm}, being well-suited for this task. However, learning from high-dimensional sequential data complicates the problem, posing a core challenge in spatial-temporal learning. This challenge is exacerbated by the quadratic computation scaling in transformers, despite their success. Approaches like dimensionality reduction~\citep{hotelling1933analysis,tipping1999probabilistic,kingma2013auto} and predictability minimisation~\citep{schmidhuber1992pm,ghahramani1994factorial} must adhere to the principle of history compression along the temporal axis, rather than compressing spatial information at each time step independently. From the perspective of information bottleneck principle~\citep{tishby2000information}, the goal is to selectively extract the ``bottleneck'' from high-dimensional sequences that is most useful for world modelling tasks like predicting future events.

Recently, the emergence of Mamba~\citep{gu2023mamba} and other SSM-based frameworks~\citep{gu2021efficiently,dao2024transformers,wang2024state} has garnered widespread attention for their strong performance in efficient sequence modelling. Mamba structure is similar to LSTM~\citep{hochreiter1997long}, in the sense that it utilises a forget gate, an input gate, and an output gate. The key difference is that these gates depend only on the previous input (not on the hidden state representing the history of inputs so far). While this hinders their representation power (e.g., cannot solve the parity problem~\citep{hochreiter1996bridging,schmidhuber2007training,srivastava2015training}), this formulation enables parallel computation of selective history compression via sub-linear sequential attention~\citep{dao2024transformers}, constructing dependencies between distant data points within the sequence. This sparks their successful applications across various tasks including language modelling~\citep{mehta2022long,grazzi2024mamba,he2024densemamba}, deep noise suppression~\citep{du2024spiking}, and clinical note understanding~\citep{yang2024clinicalmamba}. Additionally, many SSM-based vision models have been proposed for tasks such as classification~\citep{du2024understanding,shi2024vmambair,baron20232,huang2024localmamba,smith2022simplified,nguyen2022s4nd}, detection~\citep{chen2024mim}, segmentation~\citep{yang2024remamber,ma2024rs}, generation~\citep{yan2024diffusion,fei2024scalable}, and video understanding~\citep{islam2022long,wang2023selective}. Despite their success, existing SSMs often lack efficient mechanisms for handling high-dimensional spatial data, relying primarily on linear and rigid structural biases~\citep{gu2021efficiently,dao2024transformers}, particularly when dealing with permutable spatial structures.

Multivariate time-series forecasting (MTSF) and object-centric representation learning (OCRL) both involve working with noisy, high-dimensional sequential data, making them critical benchmarks for evaluating world models with significant real-world impact. We chose to assess our models on these tasks because they exemplify the primary challenge of spatial-temporal learning. Existing MTSF methods either struggle to effectively model long-term temporal dependencies~\citep{salinas2020deepar,zhang2023crossformer,Yuqietal2023patchtst} or fail to effectively leverage the cross-variate regularities in high-dimensional inputs~\citep{wu2021autoformer,zhou2022fedformer,wu2022timesnet,zeng2023dlinear,wang2023timemixer}, resulting in inaccuracies in future-state forecasting. Two notable models in the field, iTransformer~\citep{liu2023itransformer} and S-Mamba~\citep{wang2024smamba}, also have limitations. iTransformer suffers from the quadratic scaling of transformers, making it difficult to capture long-term dependencies, while S-Mamba struggles with handling the spatial structures of the data.
Object-centric representations, or ``slots'', are designed to capture ``objects'', i.e. solving the binding problem~\citep{greff2020binding}. Our goal is more general: we aim to capture modularities, or ``factors'', that remain invariant across sequences, framing the discovery of spatial regularities in history compression as another instance of the binding problem. Although a philosophical discussion on whether these ``factors'' should align with common-sense ``objects'' is beyond the scope of this paper, OCRL is closely related and serves as a good demonstration of our approach. OCRL originated from the \textit{vision-as-inverse-Bayes} framework~\citep{yuille2006vision}, initially applied to images~\citep{burgess2019monet,greff2019multi,locatello2020objectcentric}, later extended to videos~\citep{nanbo2020learning,kipf2023savi}, and developed into object-centric world models~\citep{lin2020gswm,kipf2019contrastive,wu2022slotformer,stanic2023learning}. Recent OCRL works heavily rely on the Slot Attention (SA) mechanism~\citep{locatello2020objectcentric} for object discovery, which is closely related to our routing modules. We view the SA, which also satisfies the LPE and RPI properties, as a suitable but computationally expensive alternative to~\eqref{eq:attn_rout}. In addition to MTSF and OCRL, because FACTS leverages its graph-structured memory for spatial-temporal prediction, it is also closely related to geometric modelling and learning~\citep{bronstein2017geometric,wu2020gnnsurvey}. In this paper, we demonstrate FACTS' potential for handling non-Euclidean data through long-term graph node prediction experiments.

It is worth mentioning~\citet{goyal2020object,goyal2022neural}, which propose latent state factorisation and equivariance in attention-augmented LSTM/GRU-based frameworks. However, unlike these works, FACTS employs structured state-space memory, enabling dynamic input-to-factor assignments with explicit permutation invariance and efficient training. Recently, SlotSSMs~\citep{jiang2024slot} introduced factorisation into SSMs by stacking slot attentions on top of Mamba blocks, adding an extra layer to handle modular patterns. In contrast, FACTS offers a fundamentally different state-space formulation, seamlessly integrating factorisation, geometric modelling, and selectivity within a single block through dynamic memory-input routing. While SlotSSMs can be viewed as a special case of FACTS under specific conditions (akin to the fully parallel FACTS), FACTS' versatile architecture ensures robust performance across diverse tasks and evolving environments without task- or modality-specific modifications.  



\section{Proofs} \label{append:proofs}
\begin{theorem*}[Restatement of Theorem~\ref{therm:facts}]
$\FACTS$ as defined in~\eqref{eq:pfacts_t} is L.P.E. and R.P.I. 
\end{theorem*}
\begin{proof}
Let $\sigma_Z  \in S_k$, $\sigma_X^1, \sigma_X^2, \dots, \sigma_X^t \in S_m$, $Z_0, Z_t \in \R^{k \times d}$, $X_1, X_2, \dots, X_t \in \R^{m \times d}$ .
By~\eqref{eq:pfacts_rnn}, \eqref{eq:pfacts_rr} \eqref{eq:pfacts_parallel}, and \eqref{eq:pfacts_t}, it is sufficient to show
\[
\sigma_Z \FACTS(Z_k, Z_0, X_k) = \FACTS(\sigma_Z Z_k, \sigma_Z Z_0, \sigma_X^t X_k).
\]
for all $k \in \mathbb{N}, k \in [1, t]$.
\begin{align*}
&\sigma_Z \FACTS(Z_k, Z_0, X_k)\\ 
=& \sigma_Z(\Bar A(Z_0, X_k) \odot Z_{k-1} + \Bar B(Z_0, X_k) \odot U(Z_0, X_k))\\
=& \sigma_Z \Bar A(Z_0, X_k) \odot \sigma_ZZ_{k-1} + \sigma_Z\Bar B(Z_0, X_k) \odot \sigma_ZU(Z_0, X_k).
\end{align*}
Since $\Bar \mA, \Bar \mB, \mU$ are L.P.E. and R.P.I.,
\begin{align*}
&\sigma_Z \FACTS(Z_k, Z_0, X_k)\\
=& \sigma_Z \Bar A(Z_0, X_k) \odot \sigma_ZZ_{k-1} + \sigma_Z\Bar B(Z_0, X_k) \odot \sigma_ZU(Z_0, X_k)\\
=&\Bar A(\sigma_ZZ_0, \sigma_X^kX_k) \odot \sigma_ZZ_{k-1} + \Bar B(\sigma_ZZ_0, \sigma_X^kX_k) \odot U(\sigma_ZZ_0, \sigma_X^kX_k)\\
=&\FACTS(\sigma_Z Z_k, \sigma_Z Z_0, \sigma_X^t X_k).
\end{align*}
\end{proof}
\begin{theorem*}[Restatement of Theorem~\ref{therm:generalised}]
if $\Bar \mA, \Bar \mB, \mU$ are L.P.E. and R.P.I. functions of $Z_{t-1}$ and $X_t$, any dynamics governed by~\eqref{eq:facts_ssm} is L.P.E. and R.P.I. 
\end{theorem*}
\begin{proof}
Let $Z_{t-1} \in \R^{k\times d}, X_t \in \R^{m\times d}, \sigma_Z \in S_k,$ and $\sigma_X \in S_m$ be matrices.
Assume $\Bar \mA, \Bar \mB, \mU$ are L.P.E. and R.P.I. functions of $Z_{t-1}$ and $X_t$.
By expanding~\eqref{eq:facts_ssm},
\begin{align}
\label{eq:facts_expand}
&\Bar A_t(\sigma_Z Z_{t-1}, \sigma_X X_t) \odot \sigma_Z Z_{t-1}  + \Bar B_t(\sigma_Z Z_{t-1}, \sigma_X X_t) \odot U_t(\sigma_Z Z_{t-1}, \sigma_X X_t)\\
=&\sigma_Z\Bar A_t( Z_{t-1}, X_t) \odot \sigma_Z Z_{t-1} + \sigma_Z\Bar B_t( Z_{t-1}, X_t) \odot \sigma_Z U_t( Z_{t-1},  X_t)\\
=&\sigma_Z(\Bar A_t(Z_{t-1}, X_t) \odot Z_{t-1} + \Bar B_t(Z_{t-1}, X_t) \odot Z_{t-1} \odot U_t(Z_{t-1}, X_t))\\
=& \sigma_Z Z_t 
\end{align}
\end{proof}

\section{Implementation Details} \label{append:impl}

\subsection{FACTS Module} \label{append:arch}
We provide the pseudo implementation of FACTS in Algorithm~\ref{algo:facts}, and refer to our~\faGithub~\hyperlink{https://github.com/NanboLi/FACTS}{\textcolor{palatinateblue}{Github page}} for more details. All results reported for FACTS in this paper were generated using a single NVIDIA A100 GPU (80 GB).
\begin{algorithm}[t]  
\caption{$\FACTS$ Module: a Pseudo Implementation}
\begin{algorithmic}[1]  
    \State \textbf{Input:}  $X_{1:t} \in \R^{t \times m \times d}$ 
            \Comment{$t$-sequential axis, $m$-spatial axis}
    \State \textbf{Output:} $Z_{1:t} \in \R^{t \times k \times d}$
    \State \textit{Initialise:}   $Z_0 \in \R^{k \times d}$ \Comment{E.g., init using learnables, or Gaussian samples}
    \State \textit{Param:}  $\mA=\alpha \in \R$
    \State
    \State $ X_{1:t} \gets \text{Rearrange}(\text{Linear}(X_{1:t}), [t, m, p]\rightarrow [p, t, m])$
           \Comment{Input projection}  
    \State $ (X_{1:t}, \Delta_{1:t}, \mB_{1:t}, \mC_{1:t}) \gets \text{Conv2D}(X_{1:t}; \text{ksz=(dconv, 1)})\text{.split}(\text{axis=1})$
           \Comment{Conv along $t$-axis}
    \State $ (\mU_{1:t}, \Delta_{1:t}, \mB_{1:t}, \mC_{1:t}) \gets \text{Rout}(Z_0, ({X}_{1:t}, {\Delta}_{1:t}, {\mB}_{1:t}, {\mC}_{1:t}))$
           \Comment{Routing for the elements\footnotemark}
    \State $(\Bar{\mA}_{1:t}, \Bar{\mB}_{1:t}) \gets \text{Discretisation}(\text{Softplus}(\Delta_{1:t}), \mA_{1:t}, \mB_{1:t})$
    \State $ Z_{1:t} \gets \text{StateSpacePropagation}(Z_0, \mU_{1:t}, \Bar{\mA}_{1:t}, \Bar{\mB}_{1:t})$
    \Comment{$Z_{1:t} \in \R^{t\times k \times d}$, c.f.~\eqref{eq:pfacts_rnn}-\ref{eq:pfacts_parallel}}            
    \State $\Hat{Z}_{1:t} \gets \mC_{1:t} \odot Z_{1:t}$             \Comment{$\Hat{Z}_{1:t} \in \R^{t\times k \times d}$ for selective output, e.g. $\text{Dec}(\Hat{Z}_{1:t})$}
    \State
    \State \textbf{Return:} $\Hat{Z}_{1:t}, \,Z_{1:t}$ 
          \Comment{$\Hat{Z}_{1:t}$: the output; $\,Z_{1:t}$: the state representation}
\end{algorithmic}
\label{algo:facts}
\end{algorithm}
\footnotetext{We omit the reshaping operation of the Conv2D output tensors ($[*,t,m] \rightarrow [t,m,*]$) for simplicity.}

\subsection{Long-Term Multivariate Time-Series Forecasting} \label{append:LTF}
\textbf{Datasets} We use the open-source Time Series Library (TSLib), a widely-used benchmark for training and evaluating time-series models. TSLib provides standardized settings and a leaderboard of top-performing models, ensuring fair and consistent comparisons. Our focus is on long-term multivariate time-series forecasting (MSTF) tasks. We use a collection of 9 diverse real-world datasets, as presented in Table~\ref{table:mtsf_data}. These datasets span various domains, including energy production/consumption, finance exchange rate, traffic monitoring, and weather forecasting, offering a diverse range of variates and time granularities. The temporal resolutions of the datasets range from minutes to days, making them ideal for evaluating the performance of forecasting models over both short-term and long-term horizons.

\begin{table}[htbp]
\centering
\caption{The datasets for MTSF evaluation.}
\begin{tabular}{lccc}
    \toprule
    \textbf{Datasets} & \textbf{Variates} & \textbf{Timesteps} & \textbf{Granularity} \\
    \midrule
    ETTm1 \& ETTm2     & 7    & 17,420  & 15min    \\
    ETTh1 \& ETTh2     & 7    & 69,680  & 1hour    \\
    Electricity        & 321  & 26,304  & 1hour    \\
    Traffic            & 862  & 17,544  & 1hour    \\
    Exchange           & 8    & 7,588   & 1day     \\
    Weather            & 21   & 52,696  & 10min    \\
    Solar-Energy       & 137  & 52,560  & 10min    \\
    \bottomrule
\end{tabular}
\label{table:mtsf_data}
\end{table}

\textbf{Baselines} We compare FACTS against 8 baseline models, including state-of-the-art MSTF approaches that top the TSLib leaderboard~\citep{wang2024smamba, liu2023itransformer, wu2022timesnet,Yuqietal2023patchtst,zeng2023dlinear,zhang2023crossformer,zhou2022fedformer,wu2021autoformer}. Below is a brief summary of 5 most notable baseline approaches: 
\begin{itemize}[left=1em]
\item \textit{S-Mamba}~\citep{wang2024smamba}: This baseline adapts Mamba models for MTS data by utilising a bidirectional scan on variates, achieving superior results compared to the previous leading method, iTransformer.
\item \textit{iTransformer}~\citep{liu2023itransformer}: An inverted transformer architecture that captures univariate history and cross-variate dependencies within a look-back window, though limited by the quadratic scaling of transformers. iTransformer has been leading the long-term forecasting task 
\item \textit{TimesNet}~\citep{wu2022timesnet}: Specialises in modelling multi-periodicity and interactions among periodic signals in MTS data. 
\item \textit{CrossFormer}~\citep{zhang2023crossformer}: The emphasis is on modelling cross-dimension (spatial) interactions within MTS data.
\item \textit{PatchTST}~\citep{Yuqietal2023patchtst}: Uses patching techniques to segment sub-time sequences and model channel-wise transitions, improving temporal modelling.
\end{itemize}

\textbf{FACTS for MTSF} Due to the noisy nature of raw input data, a single time step (represented as a multivariate vector) often carries limited meaningful information. A common approach to handle this is to introduce feature encoders to pre-process the data---as those temporal and positional embedders used in the baselines. We employ a set embedder to map the input multivariate sequences of size $t \times m$ into $t \times m \times d$, augmenting the tensor with an additional dimension that allows each time step to be represented as a set of $m$ features, each of $d$-dimensional size. This resulting tensor, $t \times m \times d$, serves as the direct input to the FACTS model. For prediction, we adhere to the standard practice in the Time Series Library (TSLib) benchmark, which treats time-series models as encoders designed for single-step predictions, rather than auto-regressive forecasting. We show the MTSF process of FACTS in Algorithm~\ref{algo:mtsf}.  
\begin{algorithm}  
\caption{\ac{FACTS} for Multivariate Time Series Forecasting}
\begin{algorithmic}[1]  
    \State \textbf{Input:} $\vx_{1:t} \in \R^{t \times m}$
    \State \textbf{Output:} $\vx_{t+1:t+f} \in \R^{f \times m}$
    \State
    \State $X_{1:t} \gets \text{SetEmbedder}(\vx_{1:t})$ \Comment{$X_{1:t} \in \R^{t \times m \times d}$, pre-normalisation, embedding}
    \State $Z_{1:t} \gets \FACTS(X_{1:t})$ \Comment{$Z_{1:t} \in \R^{t \times k \times d}$, $\FACTS()$ an encoder encapsulating Algo.~\ref{algo:facts}}
    \State $Z_{t+1:t+f} \gets \text{Predictor}(Z_{1:t})$ \Comment{$Z_{t+1:t+f} \in \R^{f \times k \times d}$, where $f$ is the prediction length}
    \State $\vx_{t+1:t+f} \gets \text{FactorGraphDecoder}(Z_{t+1:t+f})$ \Comment{$\vx_{t+1:t+f} \in \R^{f \times m}$}
    \State $\vx_{t+1:t+f} \gets \vx_{t+1:t+f}+ \text{ResidualPred}(\vx_{1:t})$
    \Comment{Predictive residual $\in \R^{f \times m}$ feed-through}
    \State $\vx_{t+1:t+f} \gets \text{PostProcessing}(\vx_{t+1:t+f})$
    \Comment{Invert pre-normalisation}
    \State
    \State \textbf{Return:} $\vx_{t+1:t+f}$
\end{algorithmic}
\label{algo:mtsf}
\end{algorithm}
Note that our decoder, namely ``factor-graph decoder'', is crucially designed to be invariant to the permutation of factors and processes the latent factors in parallel. Each factor independently makes predictions without relying on others. Specifically, the decoder aggregates the individual predictions of each factor by applying a softmax-weighted sum, ensuring that the final prediction effectively combines the contributions of all factors while maintaining permutation invariance. We details the aforementioned modules, i.e. the three embedders and the factor-graph decoder in the following list:
\begin{itemize}[left=1em]
\item \textbf{Discrete-Fourier Transform (DFT) decomposer}: this embedder applies the Fast Fourier Transform to each variate and decompose it into multiple spectral components. The top-$k$ low-frequency signals are treated as the ``trend component'', while the high frequency signals represent the ``seasonal component''. By concatenating the ``trend'' and ``seasonal'' components, we embed each univariate signal at each time point into a 2D (or $k+1$ dimensional, if the top-$k$ frequencies are not combined) vector representation, resulting in a set output that is compatible with \ac{FACTS}. 
\item \textbf{Conv2d embedder}: this embedder applies a convolution along the temporal axis of the input data to aggregate information from nearby time points (i.e., ``local context'', with the look-back window controlling the kernel size. We implement this using PyTorch’s standard Conv2d module, setting the kernel size to $[\text{lookback}, 1]$ and padding with zeros at the initial time steps. Although such embedders learn the input-feature mapping directly from data, making them flexible in capturing different relationships in different datasets, the choice of look-back window size is often intuitive. 
\item \textbf{Multi-scale Conv2d (MS-Conv2d) Embedder}: this embedder retains the learning flexibility of a Conv2d embedder while extending the single look-back window to multiple scales. By combining different scales, it captures features at varying granularities, making it the most robust among the three embedders and consistently delivering strong results (as shown in Table~\ref{table:ablation-embed}).
\item \textbf{Factor Graph Decoder (FGD)}: this decoder takes in the a set of predicted latents $Z_f=\{\vz^i_f\}_{i=1:k}$ (c.f. Algorithm~\ref{algo:mtsf} for its definition) and first project each $\vz^i_f \in \R^{d}$ (can run parallel) to $\mathbf{\Tilde{\alpha}}^i_f \in \R^{m}$ (the logits) and $\Tilde{\vx}^i_f \in \R^{m}$ (the prediction of the i-th factor). Then the $k$ logits, which correspond to the $k$ factor predictions, will be processed to $k$ categorical probabilities by a soft-max function: $\mathbf{\alpha}^i_f = softmax(\mathbf{\Tilde{\alpha}}^i_f, \{\mathbf{\Tilde{\alpha}}^j_f\}_{j \ne i})$. The output prediction is the weighted sum of the factor predictions w.r.t. their corresponding probabilities as: $\vx_f = \sum_i^k \mathbf{\alpha}^i_f \odot \Tilde{\vx}^i_f$, similar to the processes of spatial mixing and alpha blending in the vision and graphics communities~\citep{porter1984compositing,williams2004greedy,greff2017neural}. In vision tasks, the Spatial Broadcast Decoder (SBD,~\cite{watters2019spatial}), which shares similar properties with FGD but is more computationally expensive, is a more commonly used option.
\end{itemize}

\subsection{Object-Centric World Modelling} \label{append:ocwm}
\textbf{Benchmark} We conducted both the \textit{slot dynamics prediction} on the CLEVRER~\citep{yiclevrer} and OBJ3D~\citep{lin2020gswm} datasets, and the \textit{unsupervised object discovery} experiments on both the CLEVRER~\citep{yiclevrer} and MOVi-A~\citep{greff2021kubric} datasets. These datasets all consist of synthetic vision data and are used as standard benchmark for OCRL research.

\textbf{Slot-dynamics prediction}  For the slot dynamics prediction task, we follow the setup in~\citep{wu2022slotformer}, filtering out video clips with new objects entering the scene during the rollout period to ensure a consistent evaluation setting. The input to FACTS is the latent object representations extracted using a pre-trained object-slot encoder (SAVi,~\cite{kipf2023savi}) from video frames. That is, we consider the output of the pre-trained SAVi our data in such task -- same as~\citep{wu2022slotformer}. We extract the input latent object representations using a pre-trained object-slot encoder (SAVi~\cite{kipf2023savi}) from video frames, and train an auto-regressive roll-outer with a single FACTS layer to predict the latent representations for the next 10 frames, based on the latent codes from 6 observed frames. During testing, to ensure a fair comparison with SlotFormer, we burn-in the first 6 frames and roll out (predict) 48 frames. The predicted object representations are visualised using a pre-trained, frozen SAVi decoder (a spatial broadcast decoder,~\cite{wu2022slotformer}) to render video frames. 

\textbf{Unsupervised object discovery}
For unsupervised object discovery, we follow the fully-unconditional setting of SAVi, using unbiased slot initialisation and relying solely on RGB video frames as input, with no additional information. The primary modification is replacing SAVi's recurrent slot attention modules with FACTS. Importantly, all of the used modules (CNN vision encoders, FACTS, and decoders) \textit{end-to-end} in a single run without any supervision. We evaluated object discovery under two settings, i.e. video reconstruction and future-frame prediction. For both setting, the input number of frames are set to 6, while for prediction FACTS is asked to predict, in addition to the observed 6 frames, future 10 frames. The prediction setting is conducted with a loss function and a \textit{predictive residual design} to emphasise future frame prediction accuracy. The predicted noise will be added to the decoded prediction, i.e., in the pixel space of the future frames. A similar design is also used in our MSTF experiments, c.f. the $\text{\textbf{ResidualPred}}$ in Algorithm~\ref{algo:mtsf} (line 8). Such predictive residual designs allow FACTS to avoid handling pixel-level noise, enabling it to focus on environment dynamics more effectively. In our OCRL experiments, we define such a design as a mapping between input $t$-frame video to the pixel-wise noise of $f$ future video frames, denoted as $\text{\textbf{PixelResPredictor}}: t \times c \times h \times w \rightarrow f \times 1 \times h \times w$. As mentioned, in the future prediction experiments, we found this quite effective in capturing video dynamics as it sets free the slots (latent states) from capturing objects that are ``irrelevant'' for modelling their dynamics---i.e., the static objects. The static objects will be merged into the ``background''. In object reconstruction experiments, object reconstruction experiments, this residual predictor is turned off to ensure a fair comparison.

\subsection{Dynamic Graph Node Prediction} \label{append:graph-pred}
\textbf{Benchmark} We conducted experiments on the METR-LA traffic dataset, which is a benchmark data set for traffic prediction, capturing 207 county highway traffic sensor speed observations in Los Angeles metropolitan area. It contains traffic data collected from sensors placed on road segments, represented as a dynamic graph with nodes corresponding to sensors and edges capturing traffic correlations.  

\textbf{Setup} We follow the experimental setup of STGM and incorporate FACTS with a masking mechanism. Specifically, masking is applied to the routing processes described in~\eqref{eq:attn_rout}, which underpins the construction of selective state-space model (SSM) parameters, $\Bar{\mA}$ and $\Bar{\mB}$. FACTS captures this mask information and integrates it into state-space dynamics modelling, ensuring that only relevant historical data informs the predictions.


\section{Ablation Study} \label{ablation_app}
FACTS requires the use of a set embedder and the predefined selection of the number of factors in the state-space memory prior to training. Our ablation study aims to investigate the impact of different set embedders and the choice of the number of predefined factors on model performance.
\begin{table*}
\centering
\begin{tabular}{c|ccc|c}
        \toprule
        Pred.Len &  MS-Conv2d Emb. & Conv2d Emb. & DFT Emb. & Avg$\pm$Std.Err. \\
        \midrule
         96 
            & 0.143 & 0.144 & 0.147 & 0.145$\pm$0.002 \\
        192
            & 0.158 & 0.159 & 0.161 & 0.159$\pm$0.001 \\
        336
            & 0.171 & 0.170 & 0.171 & 0.171$\pm$0.000 \\
        720
            & 0.198 & 0.228 & 0.236 & 0.219$\pm$0.015 \\
        \bottomrule
    \end{tabular}
  \caption{Ablation: MSTF Performance of FACTS vs. different embedders (MSE$\downarrow$).}
  \label{table:ablation-embed}
  \vspace{-1em}
\end{table*}
\begin{table*}
\centering
\begin{tabular}{c|ccc|c}
        \toprule
        $k$ \textbackslash $d$
            &   1     &   128   &   512   & Avg.$\pm$Std.Err. \\
        \midrule
        1 
            & 0.349 & 0.330 & 0.328 & 0.336$\pm$0.007 \\ 
        3
            & 0.349 & 0.337 & 0.326 & 0.337$\pm$0.007 \\
        5
            & 0.349 & 0.331 & 0.349 & 0.343$\pm$0.006 \\
        7
            & 0.349 & 0.331 & 0.330 & 0.337$\pm$0.006 \\
        9
            & 0.349 & 0.332 & 0.352 & 0.344$\pm$0.006 \\
        \midrule
        Avg.$\pm$Std.Err.
            & 0.349$\pm$0.000 & 0.332$\pm$0.001 & 0.336$\pm$0.006 & ----- \\
        \bottomrule
    \end{tabular}
  \caption{Ablation: MSTF Performance of FACTS vs. (\#factors $k$, \#dimensions $d$)  (MSE$\downarrow$).}
  \label{table:ablation-kd}
  \vspace{-1em}
\end{table*}

\textbf{Impact of different set encoders} We conduct our experiments on the Electricity dataset, testing four prediction lengths: 96, 192, 336, and 720. Three different set encoders are evaluated, each employing different priors: a Discrete-Fourier Transform (DFT) decomposer, a trainable Conv2d embedder, and a multi-scale Conv2d embedder (inspired by~\cite{wu2022timesnet}). Table~\ref{table:ablation-embed} presents the comparison results, with further details of these embedders provided in Appendix~\ref{append:LTF}. The multi-scale periodic embedder consistently outperforms the others across all prediction lengths, while the DFT-based embedder shows declining performance as the prediction length increases. The standard error further indicates that longer forecasting horizons amplify the impact of encoder choice, making it a more critical factor in model accuracy. This highlights the importance of using an unbiased, learnable set encoder to improve generalisation.

\textbf{Impact of different number of factors} Previous work, such as Mamba~\citep{gu2023mamba} and xLSTM~\citep{beck2024xlstm}, shows that state-space memory size significantly impacts performance. In FACTS, memory size is determined by the number of factors and the dimension of each factor ($d$). To provide a comprehensive analysis, we examine the impact of the preset number of factors ($k$) on FACTS’ performance using the MOVi-A videos and the ETTm1 dataset. Specifically, we show the impact of $k$ during testing time in the MOVi-A video reconstruction experiments and the robustness of FACTS in training time in ETTm1 MTS forecasting experiments (with a 96-prediction length setting).

\begin{figure}[htp]
    \centering
    \begin{minipage}{0.35\textwidth}
        \centering
        \includegraphics[width=\textwidth]{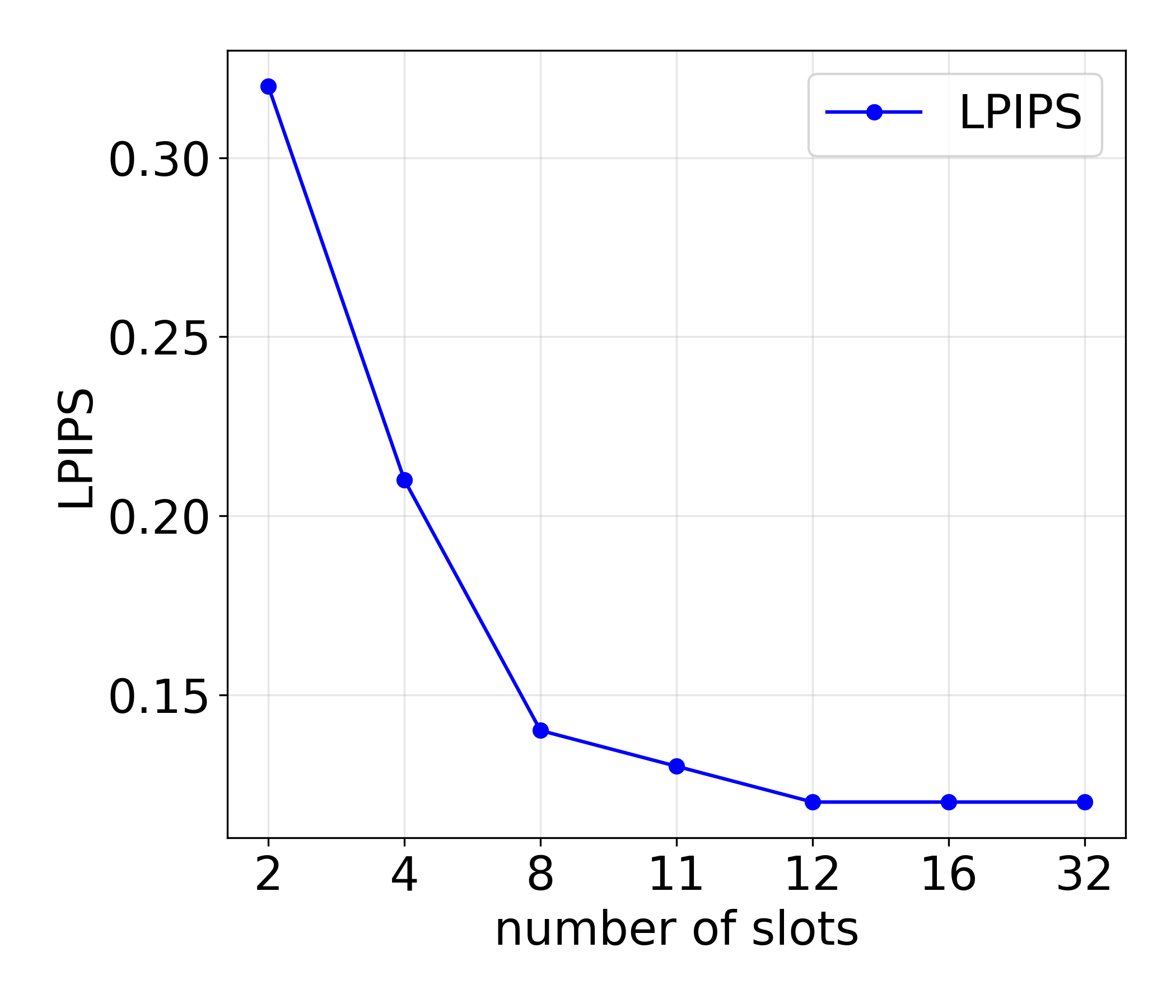} 
        \caption{Video reconstruction quality vs number of slots. With video reconstruction quality measure by LPIPS$\downarrow$.}
        \label{fig:movia-vs-k}
    \end{minipage} 
    \hfill
    \begin{minipage}{0.6\textwidth}
        \centering
        \begin{tabular}{l|ccc}
            \toprule
                \textbf{Method} & ARI $\uparrow$ & FG-ARI $\uparrow$ & FG-mIoU $\uparrow$ \\
            \midrule
                G-SWM~(\citeyear{lin2020gswm}) & 57.14 & 49.61 & 24.44 \\
                SlotFormer~(\citeyear{wu2022slotformer}) & \textbf{63.45} & 63.00 & 29.81 \\
                SAVi-dyn~(\citeyear{kipf2023savi}) & 8.64 & \textbf{64.32} & 18.25 \\
            \midrule
                \textbf{FACTS (Ours)} & 58.25 & 62.34 & \textbf{48.11} \\
            \bottomrule
        \end{tabular}
        \captionof{table}{Quantitative results for slot dynamics prediction: segmentation quality of predicted future frames, measured by ARI, FG-ARI, and FG-mIoU (all reported in \%, higher indicate better).}
        \label{table:slot-dyn-seg}
    \end{minipage}
\end{figure}

For the MOVi-A experiments, we took a FACTS model that is trained under the ``object dicscovery for video reconstruction'' setting and  evaluated its video reconstruction performance (measured by the image visual quality measure, LPIPS) against different preset $k$ at testing time on the MOVi-A data. Our results in Figure~\ref{fig:movia-vs-k} show that the video reconstruction quality can be improved by increasing $k$ up to certain level ($>11$). Knowing that the maximum number of objects in these videos (the true causal factors) is 11 (10 objects plus 1 background), suggesting a ``sweet point'' that is both effective and computationally efficient. However, in practice, identifying such a point in advance can be challenging, Figure~\ref{fig:movia-vs-k} indicates that a larger $k$ is preferred.

As the choice of $k$ could largely affect FACTS's performance, we want to investigate how robust FACTS is against different choices of $k$. We examined this on the MTS forecasting task using the ETTm1 data, which consists of 7 variates. To isolate the effect of the number of factors, we train the model across various settings, gridded by different numbers of factors and factor dimensions. As shown in Table~\ref{table:ablation-kd}, FACTS achieves consistent performance across different number of factors (during training) and also the factor dimensions, demonstrating FACTS' robustness to these hyper-parameters.

\section{Additional Results} \label{append:add_results}

\begin{table*}[h!] 
  \centering
  \renewcommand\tabcolsep{15pt}
  \renewcommand\arraystretch{1.2}
  \footnotesize
  \resizebox{\textwidth}{!}{
  \begin{tabular}{l|c|cc|cc|cc|cc|cc|cc|cc|cc|cc}
    \Xhline{1.pt}
    \multirow{2}{*}{\textbf{Data}}
        & & \multicolumn{2}{c|}{\textbf{FACTS (Ours)}} & \multicolumn{2}{c|}{S-Mamba} & \multicolumn{2}{c|}{iTransformer} & \multicolumn{2}{c|}{TimesNet} & \multicolumn{2}{c|}{PatchTST} & \multicolumn{2}{c|}{DLinear} & \multicolumn{2}{c|}{Crossformer} & \multicolumn{2}{c|}{FEDformer} & \multicolumn{2}{c}{Autoformer} \\ \cline{2-20} 
        & Pred.Len. & MSE$\downarrow$  & MAE$\downarrow$  & MSE$\downarrow$  & MAE$\downarrow$  & MSE$\downarrow$  & MAE$\downarrow$  & MSE$\downarrow$  & MAE$\downarrow$  & MSE$\downarrow$  & MAE$\downarrow$  & MSE$\downarrow$  & MAE$\downarrow$  & MSE$\downarrow$  & MAE$\downarrow$  & MSE$\downarrow$  & MAE$\downarrow$  & MSE$\downarrow$  & MAE$\downarrow$  \\ \hline
    \multirow{5}{*}{ETTm1} 
        & 96   & \run{0.330} & \win{0.364} & 0.333 & 0.368 & 0.334 & 0.368 & 0.338 & 0.375 & \win{0.329} & \run{0.367} & 0.345 & 0.372 & 0.404 & 0.426 & 0.379 & 0.419 & 0.505 & 0.475 \\ 
        & 192  & \run{0.370} & \win{0.384} & 0.376 & 0.390 & 0.377 & 0.391 & 0.374 & 0.387 & \win{0.367} & \run{0.385} & 0.380 & 0.389 & 0.450 & 0.451 & 0.426 & 0.441 & 0.553 & 0.496 \\ 
        & 336  & \run{0.400} & \win{0.404} & 0.408 & 0.413 & 0.426 & 0.420 & \run{0.410} & 0.411 & \win{0.399} & 0.410 & 0.413 & 0.413 & 0.532 & 0.515 & 0.445 & 0.459 & 0.621 & 0.537 \\ 
        & 720  & \run{0.468} & \win{0.438} & 0.475 & 0.448 & 0.491 & 0.459 & 0.478 & 0.450 & \win{0.454} & \run{0.439} & 0.474 & 0.453 & 0.666 & 0.589 & 0.543 & 0.490 & 0.671 & 0.561 \\
        \cline{2-20}
        & Avg. & \run{0.392} & \win{0.397} & 0.398 & 0.405 & 0.407 & 0.410 & 0.400 & 0.406 & \win{0.387} & \run{0.400} & 0.403 & 0.407 & 0.513 & 0.496 & 0.448 & 0.452 & 0.588 & 0.617 \\ 
    \Xhline{1.pt}
    \multirow{5}{*}{ETTm2} 
        & 96   & \win{0.175} & \win{0.258} & 0.179 & 0.263 & 0.180 & 0.264 & 0.187 & 0.267 & \win{0.175} & \run{0.259} & 0.193 & 0.292 & 0.287 & 0.366 & 0.203 & 0.287 & 0.255 & 0.339 \\ 
        & 192  & \win{0.240} & \win{0.300} & 0.250 & 0.309 & 0.250 & 0.309 & 0.249 & 0.309 & \run{0.241} & \run{0.302} & 0.284 & 0.362 & 0.414 & 0.492 & 0.269 & 0.328 & 0.281 & 0.340 \\
        & 336  & \win{0.304} & \win{0.341} & 0.312 & 0.349 & 0.311 & 0.348 & 0.321 & 0.351 & \run{0.305} & \run{0.343} & 0.369 & 0.427 & 0.597 & 0.542 & 0.325 & 0.366 & 0.339 & 0.372 \\
        & 720  & \run{0.404} & \win{0.400} & 0.411 & 0.406 & 0.412 & 0.407 & 0.408 & 0.403 & \win{0.402} & \win{0.400} & 0.554 & 0.522 & 1.730 & 1.042 & 0.421 & 0.415 & 0.433 & 0.432 \\
        \cline{2-20}
        & Avg. & \win{0.281} & \win{0.325} & 0.288 & 0.332 & 0.288 & 0.332 & 0.291 & 0.333 & \win{0.281} & \run{0.326} & 0.350 & 0.401 & 0.757 & 0.610 & 0.305 & 0.349 & 0.327 & 0.371 \\
    \Xhline{1.pt}
    \multirow{5}{*}{ETTh1} 
        & 96   & \run{0.382} & \win{0.390} & 0.386 & 0.405 & 0.386 & 0.405 & 0.384 & 0.402 & 0.414 & 0.419 & 0.386 & \run{0.400} & 0.423 & 0.448 & \win{0.376} & 0.419 & 0.449 & 0.459 \\
        & 192  & \run{0.433} & \win{0.419} & 0.443 & 0.437 & 0.441 & 0.436 & 0.436 & \run{0.429} & 0.460 & 0.445 & 0.437 & 0.432 & 0.471 & 0.474 & \win{0.420} & 0.448 & 0.500 & 0.482 \\
        & 336  & \run{0.474} & \win{0.440} & 0.489 & 0.468 & 0.487 & \run{0.458} & 0.491 & 0.469 & 0.501 & 0.466 & 0.481 & 0.459 & 0.570 & 0.546 & \win{0.459} & 0.465 & 0.521 & 0.496 \\
        & 720  & \win{0.472} & \win{0.463} & 0.502 & 0.489 & 0.503 & 0.491 & 0.521 & 0.500 & \run{0.500} & \run{0.488} & 0.519 & 0.516 & 0.653 & 0.621 & 0.506 & 0.507 & 0.514 & 0.512 \\
        \cline{2-20}
        & Avg. & \win{0.440} & \win{0.428} & 0.455 & 0.450 & 0.454 & \run{0.447} & 0.458 & 0.450 & 0.469 & 0.454 & 0.456 & 0.452 & 0.529 & 0.522 & \win{0.440} & 0.460 & 0.496 & 0.487 \\
    \Xhline{1.pt}
    \multirow{5}{*}{ETTh2} 
        & 96   & \win{0.288} & \win{0.337} & \run{0.296} & \run{0.348} & 0.297 & 0.349 & 0.340 & 0.374 & 0.302 & \run{0.348} & 0.333 & 0.387 & 0.745 & 0.584 & 0.358 & 0.397 & 0.346 & 0.388 \\ 
        & 192  & \win{0.368} & \win{0.393} & \run{0.376} & \run{0.396} & 0.380 & 0.400 & 0.402 & 0.414 & 0.388 & 0.400 & 0.477 & 0.476 & 0.877 & 0.656 & 0.429 & 0.439 & 0.456 & 0.452 \\ 
        & 336  & \win{0.414} & \win{0.427} & \run{0.424} & \run{0.431} & 0.428 & 0.432 & 0.452 & 0.452 & 0.426 & 0.433 & 0.594 & 0.541 & 1.043 & 0.731 & 0.496 & 0.487 & 0.482 & 0.486 \\ 
        & 720  & \win{0.422} & \win{0.441} & \run{0.426} & \run{0.444} & 0.427 & 0.445 & 0.462 & 0.468 & 0.431 & 0.446 & 0.831 & 0.657 & 1.104 & 0.763 & 0.463 & 0.474 & 0.515 & 0.511 \\ 
        \cline{2-20}
        & Avg. & \win{0.373} & \win{0.399} & \run{0.381} & \run{0.405} & 0.383 & 0.407 & 0.414 & 0.427 & 0.387 & 0.407 & 0.559 & 0.515 & 0.942 & 0.684 & 0.437 & 0.449 & 0.450 & 0.459 \\ 
    \Xhline{1.pt}
    \multirow{5}{*}{Electricity} 
        & 96   & \run{0.143} & \run{0.240} & \win{0.139} & \win{0.235} & 0.148 & \run{0.240} & 0.168 & 0.272 & 0.181 & 0.270 & 0.197 & 0.282 & 0.219 & 0.314 & 0.193 & 0.308 & 0.201 & 0.317 \\ 
        & 192  & \win{0.155} & \win{0.252} & \run{0.159} & 0.255 & 0.162 & \run{0.253} & 0.184 & 0.289 & 0.188 & 0.274 & 0.196 & 0.285 & 0.231 & 0.322 & 0.201 & 0.315 & 0.222 & 0.334 \\ 
        & 336  & \win{0.168} & \win{0.267} & \run{0.176} & 0.272 & 0.178 & \run{0.269} & 0.198 & 0.300 & 0.204 & 0.293 & 0.209 & 0.301 & 0.246 & 0.337 & 0.214 & 0.329 & 0.231 & 0.338 \\ 
        & 720  & \win{0.197} & \win{0.294} & \run{0.204} & \run{0.298} & 0.225 & 0.317 & 0.220 & 0.320 & 0.246 & 0.324 & 0.245 & 0.333 & 0.280 & 0.363 & 0.246 & 0.355 & 0.254 & 0.361 \\
        \cline{2-20}
        & Avg. & \win{0.166} & \win{0.263} & \run{0.170} & \run{0.265} & 0.178 & 0.270 & 0.192 & 0.295 & 0.205 & 0.290 & 0.212 & 0.300 & 0.244 & 0.334 & 0.214 & 0.327 & 0.227 & 0.338 \\ 
    \Xhline{1.pt}
    \multirow{5}{*}{Exchange} 
        & 96   & \win{0.081} & \win{0.197} & \run{0.086} & 0.207 & \run{0.086} & 0.206 & 0.107 & 0.234 & 0.088 & \run{0.205} & 0.088 & 0.218 & 0.256 & 0.367 & 0.148 & 0.278 & 0.197 & 0.323 \\ 
        & 192  & \win{0.170} & \win{0.294} & 0.182 & 0.304 & 0.177 & \run{0.299} & 0.226 & 0.344 & \run{0.176} & \run{0.299} & 0.176 & 0.315 & 0.470 & 0.509 & 0.271 & 0.315 & 0.300 & 0.369 \\ 
        & 336  & \run{0.309} & \run{0.401} & 0.332 & 0.418 & 0.331 & 0.417 & 0.367 & 0.448 & \win{0.301} & \win{0.397} & 0.313 & 0.427 & 1.268 & 0.883 & 0.460 & 0.427 & 0.509 & 0.524 \\ 
        & 720  & \win{0.808} & \win{0.677} & 0.867 & 0.703 & 0.847 & \run{0.691} & 0.964 & 0.746 & 0.901 & 0.714 & \run{0.839} & 0.695 & 1.767 & 1.068 & 1.195 & 0.695 & 1.447 & 0.941 \\
        \cline{2-20}
        & Avg. & \win{0.342} & \win{0.392} & 0.367 & 0.408 & 0.360 & \run{0.403} & 0.416 & 0.443 & 0.367 & 0.404 & \run{0.354} & 0.414 & 0.940 & 0.707 & 0.519 & 0.429 & 0.613 & 0.539 \\ 
    \Xhline{1.pt}
    \multirow{5}{*}{Traffic} 
        & 96   & 0.451 & 0.298 & \win{0.382} & \win{0.261} & \run{0.395} & \run{0.268} & 0.593 & 0.321 & 0.462 & 0.295 & 0.650 & 0.396 & 0.522 & 0.290 & 0.587 & 0.366 & 0.613 & 0.388 \\ 
        & 192  & 0.458 & 0.297 & \win{0.396} & \win{0.267} & \run{0.417} & \run{0.276} & 0.617 & 0.336 & 0.466 & 0.296 & 0.598 & 0.370 & 0.530 & 0.293 & 0.604 & 0.373 & 0.616 & 0.382 \\ 
        & 336  & 0.472 & 0.302 & \win{0.417} & \win{0.276} & \run{0.433} & \run{0.283} & 0.629 & 0.336 & 0.482 & 0.304 & 0.605 & 0.373 & 0.558 & 0.305 & 0.621 & 0.383 & 0.622 & 0.337 \\ 
        & 720  & 0.507 & 0.317 & \win{0.460} & \win{0.300} & \run{0.467} & \run{0.302} & 0.640 & 0.350 & 0.514 & 0.322 & 0.645 & 0.394 & 0.589 & 0.328 & 0.626 & 0.382 & 0.660 & 0.408 \\
        \cline{2-20}
        & Avg. & 0.472 & 0.303 & \win{0.414} & \win{0.276} & \run{0.428} & \run{0.282} & 0.620 & 0.336 & 0.481 & 0.304 & 0.625 & 0.383 & 0.550 & 0.304 & 0.610 & 0.376 & 0.628 & 0.379 \\ 
    \Xhline{1.pt}
    \multirow{5}{*}{Weather} 
        & 96   & \run{0.163} & \win{0.210} & 0.165 & \win{0.210} & 0.174 & 0.214 & 0.172 & 0.220 & 0.177 & 0.218 & 0.196 & 0.255 & \win{0.158} & 0.230 & 0.217 & 0.296 & 0.266 & 0.336 \\ 
        & 192  & 0.217 & 0.258 & \run{0.214} & \win{0.252} & 0.221 & \run{0.254} & 0.219 & 0.261 & 0.225 & 0.259 & 0.237 & 0.296 & \win{0.206} & 0.277 & 0.276 & 0.336 & 0.307 & 0.367 \\ 
        & 336  & 0.275 & 0.299 & \run{0.274} & \run{0.297} & 0.278 & \win{0.296} & 0.280 & 0.306 & 0.278 & \run{0.297} & 0.283 & 0.335 & \win{0.272} & 0.335 & 0.339 & 0.380 & 0.359 & 0.395 \\ 
        & 720  & \run{0.349} & \run{0.347} & 0.350 & \win{0.345} & 0.358 & \run{0.347} & 0.365 & 0.359 & 0.354 & 0.348 & \win{0.345} & 0.381 & 0.398 & 0.418 & 0.403 & 0.428 & 0.419 & 0.428 \\
        \cline{2-20}
        & Avg. & \win{0.251} & \run{0.278} & \win{0.251} & \win{0.276} & 0.258 & \run{0.278} & 0.259 & 0.287 & 0.259 & 0.281 & 0.265 & 0.317 & 0.259 & 0.315 & 0.309 & 0.360 & 0.338 & 0.382 \\ 
    \Xhline{1.pt}
    \multirow{5}{*}{Solar-Energy} 
        & 96   & \win{0.199} & \win{0.237} & 0.205 & 0.244 & \run{0.203} & \win{0.237} & 0.250 & 0.292 & 0.234 & 0.286 & 0.290 & 0.378 & 0.310 & 0.331 & 0.242 & 0.342 & 0.884 & 0.711 \\ 
        & 192  & 0.249 & 0.271 & \run{0.237} & \run{0.270} & \win{0.233} & \win{0.261} & 0.296 & 0.318 & 0.267 & 0.310 & 0.320 & 0.398 & 0.734 & 0.725 & 0.285 & 0.380 & 0.834 & 0.692 \\ 
        & 336  & 0.276 & \run{0.285} & \run{0.258} & 0.288 & \win{0.248} & \win{0.273} & 0.319 & 0.330 & 0.290 & 0.315 & 0.353 & 0.415 & 0.750 & 0.735 & 0.282 & 0.376 & 0.941 & 0.723 \\ 
        & 720  & 0.288 & 0.293 & \run{0.260} & \run{0.288} & \win{0.249} & \win{0.275} & 0.338 & 0.337 & 0.289 & 0.317 & 0.356 & 0.413 & 0.769 & 0.765 & 0.357 & 0.427 & 0.882 & 0.717 \\
        \cline{2-20}
        & Avg. & 0.253 & \run{0.272} & \run{0.240} & 0.273 & \win{0.233} & \win{0.262} & 0.301 & 0.319 & 0.270 & 0.307 & 0.330 & 0.401 & 0.641 & 0.639 & 0.291 & 0.381 & 0.885 & 0.711 \\ 
    \Xhline{1.pt}
  \end{tabular}
  }
    \caption{Full results for the  MTS long-term forecasting task (in MSE$\downarrow$ and MAE$\downarrow$). We compare extensive competitive models under different prediction lengths. The input sequence length is set to 96 for all the datasets above. ``Avg'' represents the average across all four prediction lengths. For each metric and each dataset, the top performance and the second best are highlighted in \textcolor{red}{\textbf{red}} and \textcolor{blue}{blue}, respectively.}
    \label{tab:full_MTS}
\end{table*}

\begin{figure}[htbp]
    \begin{center}
    \includegraphics[width=\linewidth]{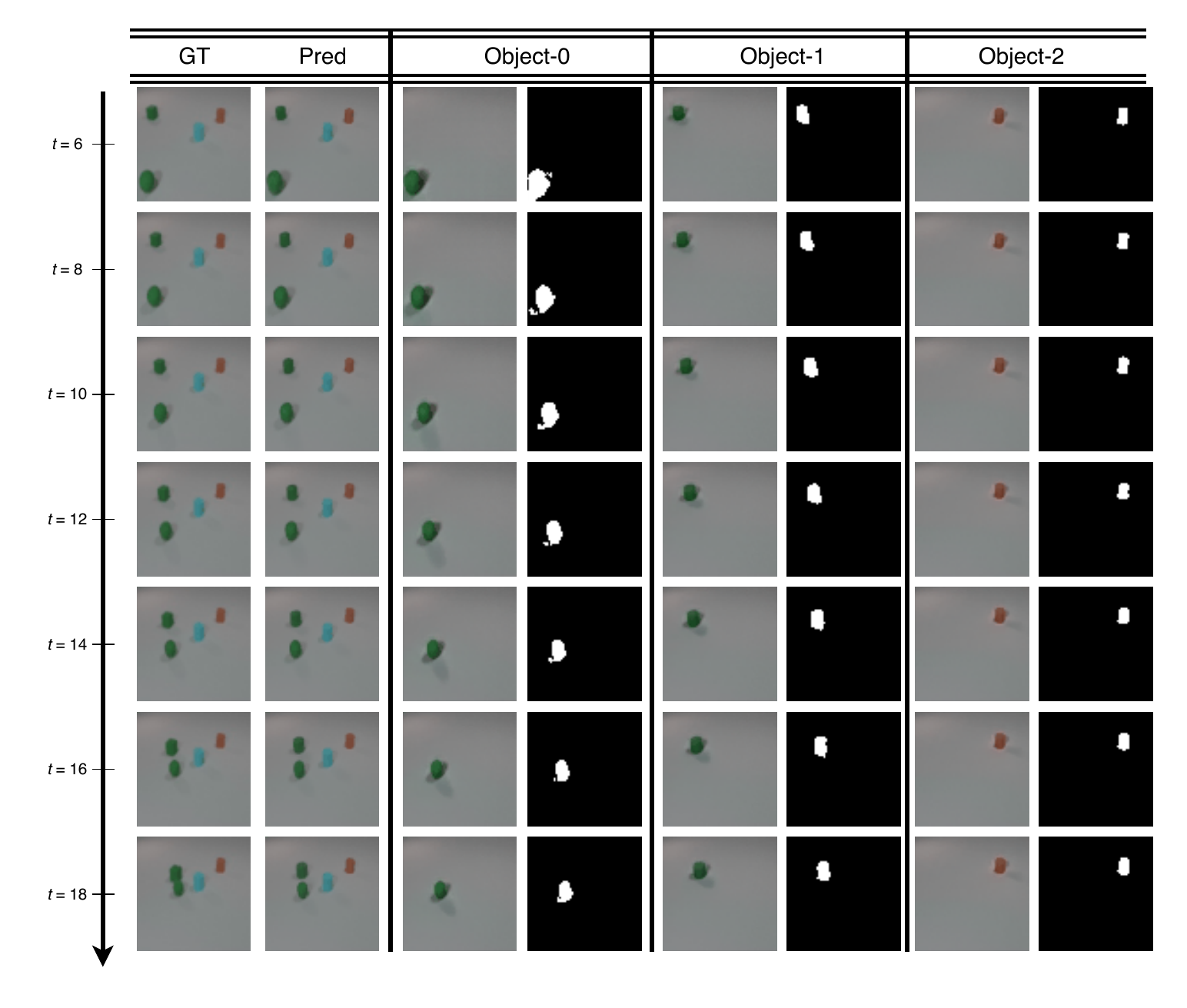} 
    \end{center}
    \caption{Qualitative results of slot dynamics prediction on CLEVRER. The first two columns show the ground truth (GT) and model predictions (Pred) for future frames. The subsequent columns represent independently rendered dynamics of individual objects (Object-0, Object-1, and Object-2) identified by the model. We show 3 object-centric dynamics in the remaining columns: two columns for each object: the left displays the predicted object dynamics, and the right shows the corresponding object masks.}
    \label{fig:facts-slots-1}
\end{figure}

\begin{figure}[htbp]
    \centering
    \begin{minipage}{\textwidth}
        \includegraphics[width=\linewidth]{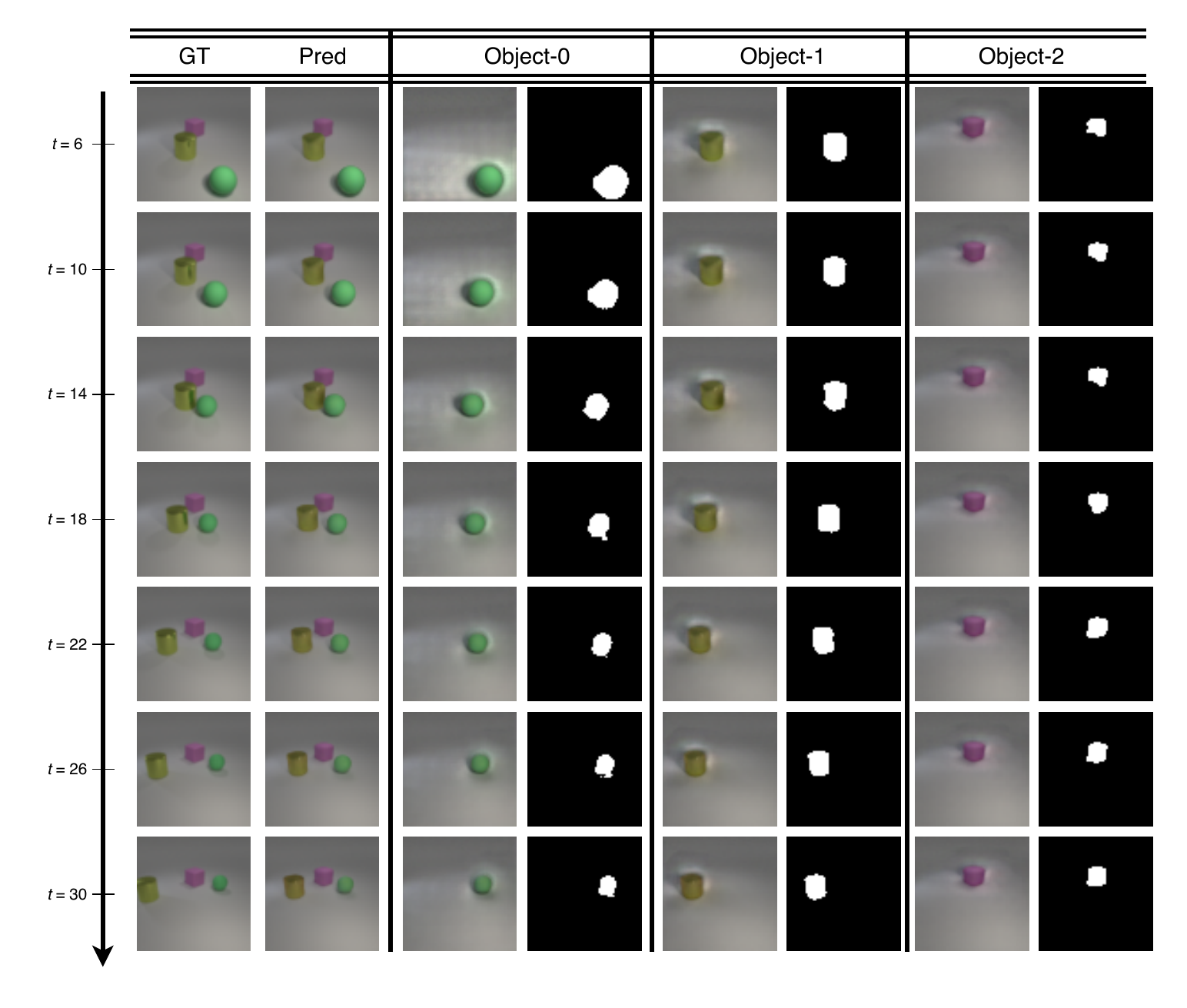} 
    \end{minipage}
    \caption{Qualitative results of slot dynamics prediction on OBJ3D. The first two columns show the ground truth (GT) and model predictions (Pred) for future frames. The subsequent columns represent independently rendered dynamics of individual objects (Object-0, Object-1, and Object-2) identified by the model. We show 3 object-centric dynamics in the remaining columns: two columns for each object: the left displays the predicted object dynamics, and the right shows the corresponding object masks.}
    \begin{minipage}{\textwidth}
        \includegraphics[width=\linewidth]{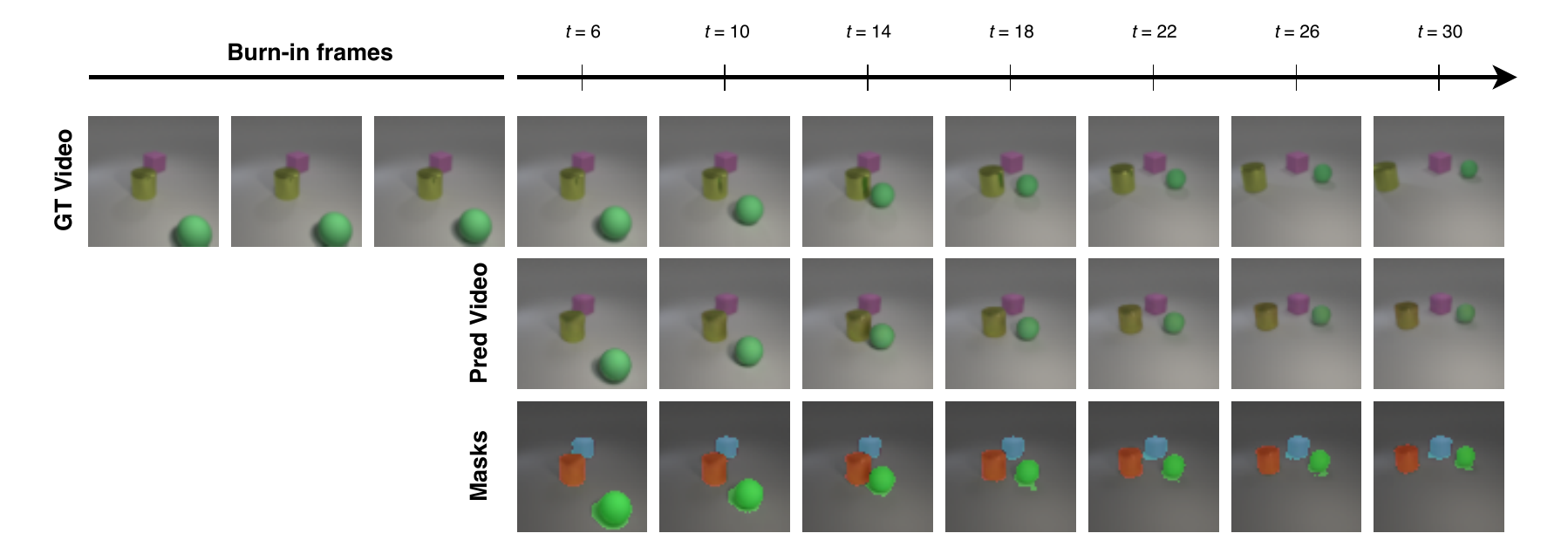} 
    \end{minipage}
    \caption{Qualitative results of slot dynamics prediction on OBJ3D. The top row shows the ground truth (GT) video frames, with burn-in frames used for initialisation. The middle row presents the predicted future frames (Pred Video) generated by the model. The bottom row illustrates the object segmentation masks predicted by the model.}
    \label{fig:facts-slots-o3d-1}
\end{figure}

\begin{figure}[htbp]
    \centering
    \includegraphics[width=\linewidth]{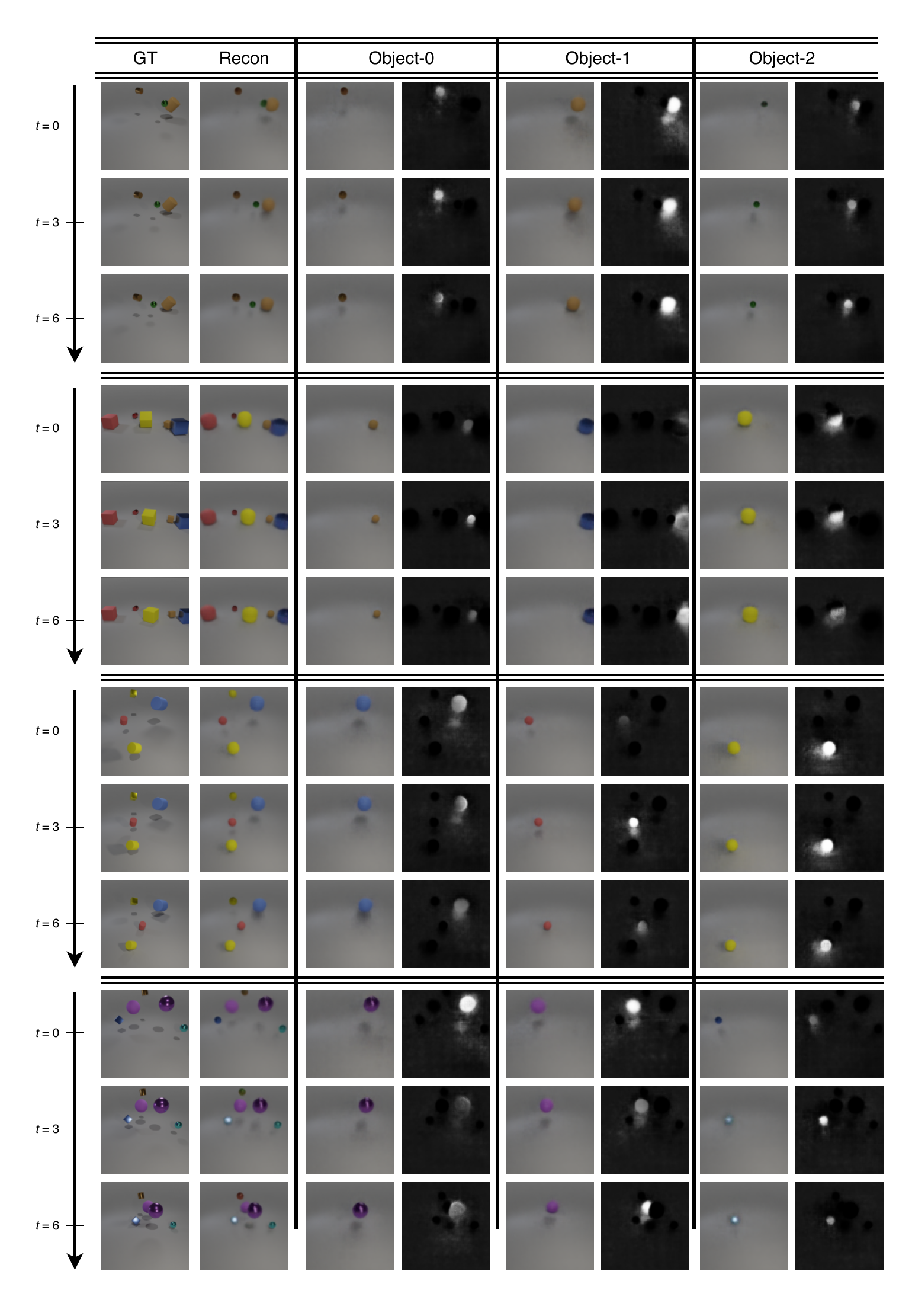} 
    \caption{Qualitative results of unsupervised object discovery. The first column shows selected ground truth future frames, while the second column presents our predicted future object-cetric states rendered as frames. We show 3 object-centric dynamics in the remaining columns: two columns for each object: the left displays the predicted object dynamics, and the right shows the corresponding attention masks. FACTS effectively discover and captures interpretable ``factors'', i.e. objects, for modelling video dynamics.}
    \label{fig:facts-unsup}
\end{figure}

\end{document}

%% file: math_commands.tex
\usepackage{amsmath,amsfonts,bm}

\def\eqref#1{equation~\ref{#1}}

\def\1{\bm{1}}

\def\vh{{\bm{h}}}

\def\vx{{\bm{x}}}
\def\vy{{\bm{y}}}
\def\vz{{\bm{z}}}

\def\mA{{\bm{A}}}
\def\mB{{\bm{B}}}
\def\mC{{\bm{C}}}
\def\mD{{\bm{D}}}

\def\mI{{\bm{I}}}

\def\mM{{\bm{M}}}

\def\mU{{\bm{U}}}

\DeclareMathAlphabet{\mathsfit}{\encodingdefault}{\sfdefault}{m}{sl}
\SetMathAlphabet{\mathsfit}{bold}{\encodingdefault}{\sfdefault}{bx}{n}

\newcommand{\R}{\mathbb{R}}

\newcommand{\softmax}{\mathrm{softmax}}



%% file: main.bbl
\begin{thebibliography}{129}
\providecommand{\natexlab}[1]{#1}
\providecommand{\url}[1]{\texttt{#1}}
\expandafter\ifx\csname urlstyle\endcsname\relax
  \providecommand{\doi}[1]{doi: #1}\else
  \providecommand{\doi}{doi: \begingroup \urlstyle{rm}\Url}\fi

\bibitem[Bai et~al.(2020)Bai, Yao, Li, Wang, and Wang]{bai2020adaptive}
Lei Bai, Lina Yao, Can Li, Xianzhi Wang, and Can Wang.
\newblock Adaptive graph convolutional recurrent network for traffic forecasting.
\newblock \emph{Advances in neural information processing systems}, 33:\penalty0 17804--17815, 2020.

\bibitem[Baron et~al.(2023)Baron, Zimerman, and Wolf]{baron20232}
Ethan Baron, Itamar Zimerman, and Lior Wolf.
\newblock A 2-dimensional state space layer for spatial inductive bias.
\newblock In \emph{The Twelfth International Conference on Learning Representations}, 2023.

\bibitem[Beck et~al.(2024)Beck, P{\"o}ppel, Spanring, Auer, Prudnikova, Kopp, Klambauer, Brandstetter, and Hochreiter]{beck2024xlstm}
Maximilian Beck, Korbinian P{\"o}ppel, Markus Spanring, Andreas Auer, Oleksandra Prudnikova, Michael Kopp, G{\"u}nter Klambauer, Johannes Brandstetter, and Sepp Hochreiter.
\newblock {xLSTM}: Extended long short-term memory.
\newblock \emph{arXiv preprint arXiv:2405.04517}, 2024.

\bibitem[Blelloch(1990)]{blelloch1990prefix}
Guy~E Blelloch.
\newblock Prefix sums and their applications.
\newblock 1990.

\bibitem[Bronstein et~al.(2017)Bronstein, Bruna, LeCun, Szlam, and Vandergheynst]{bronstein2017geometric}
Michael~M Bronstein, Joan Bruna, Yann LeCun, Arthur Szlam, and Pierre Vandergheynst.
\newblock Geometric deep learning: going beyond euclidean data.
\newblock \emph{IEEE Signal Processing Magazine}, 34\penalty0 (4):\penalty0 18--42, 2017.

\bibitem[Burgess et~al.(2019)Burgess, Matthey, Watters, Kabra, Higgins, Botvinick, and Lerchner]{burgess2019monet}
Christopher~P Burgess, Loic Matthey, Nicholas Watters, Rishabh Kabra, Irina Higgins, Matt Botvinick, and Alexander Lerchner.
\newblock {MONet}: Unsupervised scene decomposition and representation.
\newblock \emph{arXiv preprint arXiv:1901.11390}, 2019.

\bibitem[Chen et~al.(2022)Chen, Wu, Yoon, and Ahn]{chen2022transdreamer}
Chang Chen, Yi-Fu Wu, Jaesik Yoon, and Sungjin Ahn.
\newblock Transdreamer: Reinforcement learning with transformer world models.
\newblock \emph{arXiv preprint arXiv:2202.09481}, 2022.

\bibitem[Chen et~al.(2024)Chen, Tan, Gong, Chu, Wu, Liu, Ye, and Yu]{chen2024mim}
Tianxiang Chen, Zhentao Tan, Tao Gong, Qi~Chu, Yue Wu, Bin Liu, Jieping Ye, and Nenghai Yu.
\newblock {MiM-ISTD}: {Mamba}-in-{Mamba} for efficient infrared small target detection.
\newblock \emph{arXiv preprint arXiv:2403.02148}, 2024.

\bibitem[Chiappa et~al.(2017)Chiappa, Racaniere, Wierstra, and Mohamed]{chiappa2017recurrent}
Silvia Chiappa, S{\'e}bastien Racaniere, Daan Wierstra, and Shakir Mohamed.
\newblock Recurrent environment simulators.
\newblock \emph{arXiv preprint arXiv:1704.02254}, 2017.

\bibitem[Cho(2014)]{cho2014gru}
Kyunghyun Cho.
\newblock Learning phrase representations using {RNN} encoder-decoder for statistical machine translation.
\newblock \emph{arXiv preprint arXiv:1406.1078}, 2014.

\bibitem[Dao \& Gu(2024)Dao and Gu]{dao2024transformers}
Tri Dao and Albert Gu.
\newblock Transformers are {SSMs}: Generalized models and efficient algorithms through structured state space duality.
\newblock \emph{arXiv preprint arXiv:2405.21060}, 2024.

\bibitem[Deisenroth \& Rasmussen(2011)Deisenroth and Rasmussen]{deisenroth2011pilco}
Marc Deisenroth and Carl~E Rasmussen.
\newblock {PILCO}: A model-based and data-efficient approach to policy search.
\newblock In \emph{International Conference on machine learning}, 2011.

\bibitem[Denton et~al.(2017)]{denton2017unsupervised}
Emily~L Denton et~al.
\newblock Unsupervised learning of disentangled representations from video.
\newblock \emph{Advances in neural information processing systems}, 30, 2017.

\bibitem[Depeweg et~al.(2016)Depeweg, Hern{\'a}ndez-Lobato, Doshi-Velez, and Udluft]{depeweg2016learning}
Stefan Depeweg, Jos{\'e}~Miguel Hern{\'a}ndez-Lobato, Finale Doshi-Velez, and Steffen Udluft.
\newblock Learning and policy search in stochastic dynamical systems with {Bayesian} neural networks.
\newblock \emph{arXiv preprint arXiv:1605.07127}, 2016.

\bibitem[Du et~al.(2024{\natexlab{a}})Du, Li, and Xu]{du2024understanding}
Chengbin Du, Yanxi Li, and Chang Xu.
\newblock Understanding robustness of visual state space models for image classification.
\newblock \emph{arXiv preprint arXiv:2403.10935}, 2024{\natexlab{a}}.

\bibitem[Du et~al.(2024{\natexlab{b}})Du, Liu, and Chua]{du2024spiking}
Yu~Du, Xu~Liu, and Yansong Chua.
\newblock Spiking structured state space model for monaural speech enhancement.
\newblock In \emph{ICASSP 2024-2024 IEEE International Conference on Acoustics, Speech and Signal Processing (ICASSP)}, pp.\  766--770. IEEE, 2024{\natexlab{b}}.

\bibitem[Fei et~al.(2024)Fei, Fan, Yu, and Huang]{fei2024scalable}
Zhengcong Fei, Mingyuan Fan, Changqian Yu, and Junshi Huang.
\newblock Scalable diffusion models with state space backbone.
\newblock \emph{arXiv preprint arXiv:2402.05608}, 2024.

\bibitem[Finn et~al.(2016)Finn, Tan, Duan, Darrell, Levine, and Abbeel]{finn2016deep}
Chelsea Finn, Xin~Yu Tan, Yan Duan, Trevor Darrell, Sergey Levine, and Pieter Abbeel.
\newblock Deep spatial autoencoders for visuomotor learning.
\newblock In \emph{2016 IEEE International Conference on Robotics and Automation (ICRA)}, pp.\  512--519. IEEE, 2016.

\bibitem[Gal et~al.(2016)Gal, McAllister, and Rasmussen]{gal2016improving}
Yarin Gal, Rowan McAllister, and Carl~Edward Rasmussen.
\newblock Improving {PILCO} with {Bayesian} neural network dynamics models.
\newblock In \emph{Data-efficient machine learning workshop, ICML}, volume~4, pp.\ ~25, 2016.

\bibitem[Ghahramani(1994)]{ghahramani1994factorial}
Zoubin Ghahramani.
\newblock Factorial learning and the {EM} algorithm.
\newblock \emph{Advances in neural information processing systems}, 7, 1994.

\bibitem[Goyal et~al.(2020)Goyal, Lamb, Gampa, Beaudoin, Levine, Blundell, Bengio, and Mozer]{goyal2020object}
Anirudh Goyal, Alex Lamb, Phanideep Gampa, Philippe Beaudoin, Sergey Levine, Charles Blundell, Yoshua Bengio, and Michael Mozer.
\newblock Object files and schemata: Factorizing declarative and procedural knowledge in dynamical systems.
\newblock \emph{arXiv preprint arXiv:2006.16225}, 2020.

\bibitem[Goyal et~al.(2022)Goyal, Didolkar, Ke, Blundell, Beaudoin, Heess, Mozer, and Bengio]{goyal2022neural}
Anirudh Goyal, Aniket Didolkar, Nan~Rosemary Ke, Charles Blundell, Philippe Beaudoin, Nicolas Heess, Michael Mozer, and Yoshua Bengio.
\newblock Neural production systems: Learning rule-governed visual dynamics.
\newblock \emph{CoRR, abs/2103.01937}, 2022.

\bibitem[Grazzi et~al.(2024)Grazzi, Siems, Schrodi, Brox, and Hutter]{grazzi2024mamba}
Riccardo Grazzi, Julien Siems, Simon Schrodi, Thomas Brox, and Frank Hutter.
\newblock Is {Mamba} capable of in-context learning?
\newblock \emph{arXiv preprint arXiv:2402.03170}, 2024.

\bibitem[Greff et~al.(2017)Greff, Van~Steenkiste, and Schmidhuber]{greff2017neural}
Klaus Greff, Sjoerd Van~Steenkiste, and J{\"u}rgen Schmidhuber.
\newblock Neural expectation maximization.
\newblock In \emph{Advances in Neural Information Processing Systems}, pp.\  6691--6701, 2017.

\bibitem[Greff et~al.(2019)Greff, Kaufman, Kabra, Watters, Burgess, Zoran, Matthey, Botvinick, and Lerchner]{greff2019multi}
Klaus Greff, Rapha{\"e}l~Lopez Kaufman, Rishabh Kabra, Nick Watters, Christopher Burgess, Daniel Zoran, Loic Matthey, Matthew Botvinick, and Alexander Lerchner.
\newblock Multi-object representation learning with iterative variational inference.
\newblock In \emph{Proceedings of the 36th International Conference on Machine Learning}, pp.\  2424--2433, 2019.

\bibitem[Greff et~al.(2020)Greff, Van~Steenkiste, and Schmidhuber]{greff2020binding}
Klaus Greff, Sjoerd Van~Steenkiste, and J{\"u}rgen Schmidhuber.
\newblock On the binding problem in artificial neural networks.
\newblock \emph{arXiv preprint arXiv:2012.05208}, 2020.

\bibitem[Greff et~al.(2022)Greff, Belletti, Beyer, Doersch, Du, Duckworth, Fleet, Gnanapragasam, Golemo, Herrmann, Kipf, Kundu, Lagun, Laradji, Liu, Meyer, Miao, Nowrouzezahrai, Oztireli, Pot, Radwan, Rebain, Sabour, Sajjadi, Sela, Sitzmann, Stone, Sun, Vora, Wang, Wu, Yi, Zhong, and Tagliasacchi]{greff2021kubric}
Klaus Greff, Francois Belletti, Lucas Beyer, Carl Doersch, Yilun Du, Daniel Duckworth, David~J Fleet, Dan Gnanapragasam, Florian Golemo, Charles Herrmann, Thomas Kipf, Abhijit Kundu, Dmitry Lagun, Issam Laradji, Hsueh-Ti~(Derek) Liu, Henning Meyer, Yishu Miao, Derek Nowrouzezahrai, Cengiz Oztireli, Etienne Pot, Noha Radwan, Daniel Rebain, Sara Sabour, Mehdi S.~M. Sajjadi, Matan Sela, Vincent Sitzmann, Austin Stone, Deqing Sun, Suhani Vora, Ziyu Wang, Tianhao Wu, Kwang~Moo Yi, Fangcheng Zhong, and Andrea Tagliasacchi.
\newblock Kubric: a scalable dataset generator.
\newblock 2022.

\bibitem[Gu \& Dao(2023)Gu and Dao]{gu2023mamba}
Albert Gu and Tri Dao.
\newblock Mamba: Linear-time sequence modeling with selective state spaces.
\newblock \emph{arXiv preprint arXiv:2312.00752}, 2023.

\bibitem[Gu et~al.(2021)Gu, Goel, and R{\'e}]{gu2021efficiently}
Albert Gu, Karan Goel, and Christopher R{\'e}.
\newblock Efficiently modeling long sequences with structured state spaces.
\newblock \emph{arXiv preprint arXiv:2111.00396}, 2021.

\bibitem[Guo et~al.(2019)Guo, Lin, Feng, Song, and Wan]{guo2019attention}
Shengnan Guo, Youfang Lin, Ning Feng, Chao Song, and Huaiyu Wan.
\newblock Attention based spatial-temporal graph convolutional networks for traffic flow forecasting.
\newblock In \emph{Proceedings of the AAAI conference on artificial intelligence}, volume~33, pp.\  922--929, 2019.

\bibitem[Gupta et~al.(2022{\natexlab{a}})Gupta, Gu, and Berant]{gupta2022diagonal}
Ankit Gupta, Albert Gu, and Jonathan Berant.
\newblock Diagonal state spaces are as effective as structured state spaces.
\newblock \emph{Advances in Neural Information Processing Systems}, 35:\penalty0 22982--22994, 2022{\natexlab{a}}.

\bibitem[Gupta et~al.(2022{\natexlab{b}})Gupta, Mehta, and Berant]{gupta2022simplifying}
Ankit Gupta, Harsh Mehta, and Jonathan Berant.
\newblock Simplifying and understanding state space models with diagonal linear {RNNs}.
\newblock \emph{arXiv preprint arXiv:2212.00768}, 2022{\natexlab{b}}.

\bibitem[Ha \& Schmidhuber(2018)Ha and Schmidhuber]{ha2018world}
David Ha and J{\"u}rgen Schmidhuber.
\newblock World models.
\newblock \emph{Preprint arXiv:1803.10122 (variant at NeurIPS 2018)}, 2018.

\bibitem[Hafner et~al.(2019)Hafner, Lillicrap, Fischer, Villegas, Ha, Lee, and Davidson]{hafner2019learning}
Danijar Hafner, Timothy Lillicrap, Ian Fischer, Ruben Villegas, David Ha, Honglak Lee, and James Davidson.
\newblock Learning latent dynamics for planning from pixels.
\newblock In \emph{International conference on machine learning}, pp.\  2555--2565. PMLR, 2019.

\bibitem[Hafner et~al.(2020)Hafner, Lillicrap, Norouzi, and Ba]{hafner2020mastering}
Danijar Hafner, Timothy Lillicrap, Mohammad Norouzi, and Jimmy Ba.
\newblock Mastering {Atari} with discrete world models.
\newblock \emph{arXiv preprint arXiv:2010.02193}, 2020.

\bibitem[Hafner et~al.(2023)Hafner, Pasukonis, Ba, and Lillicrap]{hafner2023mastering}
Danijar Hafner, Jurgis Pasukonis, Jimmy Ba, and Timothy Lillicrap.
\newblock Mastering diverse domains through world models.
\newblock \emph{arXiv preprint arXiv:2301.04104}, 2023.

\bibitem[He et~al.(2024)He, Han, Tang, Wang, Yang, Guo, and Wang]{he2024densemamba}
Wei He, Kai Han, Yehui Tang, Chengcheng Wang, Yujie Yang, Tianyu Guo, and Yunhe Wang.
\newblock {DenseMamba}: State space models with dense hidden connection for efficient large language models.
\newblock \emph{arXiv preprint arXiv:2403.00818}, 2024.

\bibitem[Hein et~al.(2017)Hein, Depeweg, Tokic, Udluft, Hentschel, Runkler, and Sterzing]{hein2017benchmark}
Daniel Hein, Stefan Depeweg, Michel Tokic, Steffen Udluft, Alexander Hentschel, Thomas~A Runkler, and Volkmar Sterzing.
\newblock A benchmark environment motivated by industrial control problems.
\newblock In \emph{2017 IEEE Symposium Series on Computational Intelligence (SSCI)}, pp.\  1--8. IEEE, 2017.

\bibitem[Hochreiter(1991)]{hochreiter1991untersuchungen}
Sepp Hochreiter.
\newblock Untersuchungen zu dynamischen neuronalen {Netzen}.
\newblock \emph{Diploma, Technische Universit{\"a}t M{\"u}nchen}, 91\penalty0 (1):\penalty0 31, 1991.

\bibitem[Hochreiter \& Schmidhuber(1996)Hochreiter and Schmidhuber]{hochreiter1996bridging}
Sepp Hochreiter and J{\"u}rgen Schmidhuber.
\newblock Bridging long time lags by weight guessing and “long short-term memory”.
\newblock \emph{Spatiotemporal models in biological and artificial systems}, 37\penalty0 (65-72):\penalty0 11, 1996.

\bibitem[Hochreiter \& Schmidhuber(1997)Hochreiter and Schmidhuber]{hochreiter1997long}
Sepp Hochreiter and J{\"u}rgen Schmidhuber.
\newblock {Long Short-Term Memory}.
\newblock \emph{Neural Computation}, 9\penalty0 (8):\penalty0 1735--1780, 1997.

\bibitem[Hochreiter et~al.(2001)Hochreiter, Bengio, Frasconi, and Schmidhuber]{hochreiter2001gradient}
Sepp Hochreiter, Yoshua Bengio, Paolo Frasconi, and J{\"u}rgen Schmidhuber.
\newblock Gradient flow in recurrent nets: the difficulty of learning long-term dependencies, 2001.

\bibitem[Hotelling(1933)]{hotelling1933analysis}
Harold Hotelling.
\newblock Analysis of a complex of statistical variables into principal components.
\newblock \emph{Journal of educational psychology}, 24\penalty0 (6):\penalty0 417, 1933.

\bibitem[Huang et~al.(2024)Huang, Pei, You, Wang, Qian, and Xu]{huang2024localmamba}
Tao Huang, Xiaohuan Pei, Shan You, Fei Wang, Chen Qian, and Chang Xu.
\newblock {LocalMamba}: Visual state space model with windowed selective scan.
\newblock \emph{arXiv preprint arXiv:2403.09338}, 2024.

\bibitem[Islam \& Bertasius(2022)Islam and Bertasius]{islam2022long}
Md~Mohaiminul Islam and Gedas Bertasius.
\newblock Long movie clip classification with state-space video models.
\newblock In \emph{European Conference on Computer Vision}, pp.\  87--104. Springer, 2022.

\bibitem[Jiang et~al.(2024)Jiang, Deng, Singh, Lee, and Ahn]{jiang2024slot}
Jindong Jiang, Fei Deng, Gautam Singh, Minseung Lee, and Sungjin Ahn.
\newblock Slot state space models.
\newblock In \emph{Neural Information Processing Systems}, 2024.

\bibitem[Jiang et~al.(2023)Jiang, Wang, Yong, Jeph, Chen, Kobayashi, Song, Fukushima, and Suzumura]{jiang2023spatio}
Renhe Jiang, Zhaonan Wang, Jiawei Yong, Puneet Jeph, Quanjun Chen, Yasumasa Kobayashi, Xuan Song, Shintaro Fukushima, and Toyotaro Suzumura.
\newblock Spatio-temporal meta-graph learning for traffic forecasting.
\newblock In \emph{Proceedings of the AAAI conference on artificial intelligence}, volume~37, pp.\  8078--8086, 2023.

\bibitem[Kalman(1960)]{kalman1960new}
Rudolph~Emil Kalman.
\newblock A new approach to linear filtering and prediction problems.
\newblock \emph{Journal of Basic Engineering}, 82\penalty0 (1):\penalty0 35--45, 1960.

\bibitem[Kingma(2013)]{kingma2013auto}
Diederik~P Kingma.
\newblock Auto-encoding variational bayes.
\newblock \emph{arXiv preprint arXiv:1312.6114}, 2013.

\bibitem[Kipf et~al.(2019)Kipf, Van~der Pol, and Welling]{kipf2019contrastive}
Thomas Kipf, Elise Van~der Pol, and Max Welling.
\newblock Contrastive learning of structured world models.
\newblock In \emph{International Conference on Learning Representations}, 2019.

\bibitem[Kipf et~al.(2023)Kipf, Elsayed, Mahendran, Stone, Sabour, Heigold, Jonschkowski, Dosovitskiy, and Greff]{kipf2023savi}
Thomas Kipf, Gamaleldin~Fathy Elsayed, Aravindh Mahendran, Austin Stone, Sara Sabour, Georg Heigold, Rico Jonschkowski, Alexey Dosovitskiy, and Klaus Greff.
\newblock Conditional object-centric learning from video.
\newblock In \emph{International Conference on Learning Representations}, 2023.

\bibitem[Kononenko(1989)]{kononenko1989bayesian}
Igor Kononenko.
\newblock {Bayesian} neural networks.
\newblock \emph{Biological Cybernetics}, 61\penalty0 (5):\penalty0 361--370, 1989.

\bibitem[Lablack \& Shen(2023)Lablack and Shen]{lablack2023spatio}
Mourad Lablack and Yanming Shen.
\newblock Spatio-temporal graph mixformer for traffic forecasting.
\newblock \emph{Expert systems with applications}, 228:\penalty0 120281, 2023.

\bibitem[Lai et~al.(2018)Lai, Chang, Yang, and Liu]{lai2018modeling}
Guokun Lai, Wei-Cheng Chang, Yiming Yang, and Hanxiao Liu.
\newblock Modeling long-and short-term temporal patterns with deep neural networks.
\newblock In \emph{The 41st international ACM SIGIR conference on research \& development in information retrieval}, pp.\  95--104, 2018.

\bibitem[Lee \& Ko(2024)Lee and Ko]{lee2024testam}
Hyunwook Lee and Sungahn Ko.
\newblock {TESTAM}: A time-enhanced spatio-temporal attention model with mixture of experts.
\newblock In \emph{The Twelfth International Conference on Learning Representations}, 2024.
\newblock URL \url{https://openreview.net/forum?id=N0nTk5BSvO}.

\bibitem[Li et~al.(2023)Li, Feng, Yan, Jin, Yang, Sun, Jin, and Li]{li2023dynamic}
Fuxian Li, Jie Feng, Huan Yan, Guangyin Jin, Fan Yang, Funing Sun, Depeng Jin, and Yong Li.
\newblock Dynamic graph convolutional recurrent network for traffic prediction: Benchmark and solution.
\newblock \emph{ACM Transactions on Knowledge Discovery from Data}, 17\penalty0 (1):\penalty0 1--21, 2023.

\bibitem[Li et~al.(2018)Li, Yu, Shahabi, and Liu]{li2018diffusion}
Yaguang Li, Rose Yu, Cyrus Shahabi, and Yan Liu.
\newblock Diffusion convolutional recurrent neural network: Data-driven traffic forecasting.
\newblock In \emph{International Conference on Learning Representations}, 2018.

\bibitem[Lin et~al.(2020)Lin, Wu, Peri, Fu, Jiang, and Ahn]{lin2020gswm}
Zhixuan Lin, Yi-Fu Wu, Skand Peri, Bofeng Fu, Jindong Jiang, and Sungjin Ahn.
\newblock Improving generative imagination in object-centric world models.
\newblock In \emph{International Conference on Machine Learning}, pp.\  6140--6149. PMLR, 2020.

\bibitem[Liu et~al.(2024)Liu, Hu, Zhang, Wu, Wang, Ma, and Long]{liu2023itransformer}
Yong Liu, Tengge Hu, Haoran Zhang, Haixu Wu, Shiyu Wang, Lintao Ma, and Mingsheng Long.
\newblock {iTransformer}: Inverted transformers are effective for time series forecasting.
\newblock In \emph{International Conference on Learning Representations}, 2024.

\bibitem[Locatello et~al.(2020)Locatello, Weissenborn, Unterthiner, Mahendran, Heigold, Uszkoreit, Dosovitskiy, and Kipf]{locatello2020objectcentric}
Francesco Locatello, Dirk Weissenborn, Thomas Unterthiner, Aravindh Mahendran, Georg Heigold, Jakob Uszkoreit, Alexey Dosovitskiy, and Thomas Kipf.
\newblock Object-centric learning with slot attention.
\newblock In \emph{Advances in Neural Information Processing Systems}, 2020.

\bibitem[Ma et~al.(2024)Ma, Zhang, and Pun]{ma2024rs}
Xianping Ma, Xiaokang Zhang, and Man-On Pun.
\newblock {RS 3 Mamba}: Visual state space model for remote sensing image semantic segmentation.
\newblock \emph{IEEE Geoscience and Remote Sensing Letters}, 2024.

\bibitem[MacKay et~al.(1998)]{mackay1998introduction}
David~JC MacKay et~al.
\newblock Introduction to {Gaussian} processes.
\newblock \emph{NATO ASI series F computer and systems sciences}, 168:\penalty0 133--166, 1998.

\bibitem[McAllister \& Rasmussen(2016)McAllister and Rasmussen]{mcallister2016data}
Rowan McAllister and Carl~Edward Rasmussen.
\newblock Data-efficient reinforcement learning in continuous-state {POMDPs}.
\newblock \emph{arXiv preprint arXiv:1602.02523}, 2016.

\bibitem[Mehta et~al.(2022)Mehta, Gupta, Cutkosky, and Neyshabur]{mehta2022long}
Harsh Mehta, Ankit Gupta, Ashok Cutkosky, and Behnam Neyshabur.
\newblock Long range language modeling via gated state spaces.
\newblock \emph{arXiv preprint arXiv:2206.13947}, 2022.

\bibitem[Micheli et~al.(2022)Micheli, Alonso, and Fleuret]{micheli2022transformers}
Vincent Micheli, Eloi Alonso, and Fran{\c{c}}ois Fleuret.
\newblock Transformers are sample-efficient world models.
\newblock \emph{arXiv preprint arXiv:2209.00588}, 2022.

\bibitem[Munro(1987)]{munro1987dual}
Paul Munro.
\newblock A dual back-propagation scheme for scalar reward learning.
\newblock In \emph{Ninth Annual Conference of the Cognitive Science Society}, pp.\  165--176. Hillsdale, NJ. Cognitive Science Society Lawrence Erlbaum, 1987.

\bibitem[Nanbo et~al.(2020)Nanbo, Eastwood, and Fisher]{nanbo2020learning}
Li~Nanbo, Cian Eastwood, and Robert Fisher.
\newblock Learning object-centric representations of multi-object scenes from multiple views.
\newblock In \emph{Advances in Neural Information Processing Systems}, 2020.

\bibitem[Nguyen \& Widrow(1990)Nguyen and Widrow]{nguyen1990truck}
Derrick Nguyen and Bernard Widrow.
\newblock The truck backer-upper: An example of self-learning in neural networks.
\newblock In \emph{Advanced neural computers}, pp.\  11--19. Elsevier, 1990.

\bibitem[Nguyen et~al.(2022)Nguyen, Goel, Gu, Downs, Shah, Dao, Baccus, and R{\'e}]{nguyen2022s4nd}
Eric Nguyen, Karan Goel, Albert Gu, Gordon Downs, Preey Shah, Tri Dao, Stephen Baccus, and Christopher R{\'e}.
\newblock {S4ND}: Modeling images and videos as multidimensional signals with state spaces.
\newblock \emph{Advances in neural information processing systems}, 35:\penalty0 2846--2861, 2022.

\bibitem[Nie et~al.(2023)Nie, H.~Nguyen, Sinthong, and Kalagnanam]{Yuqietal2023patchtst}
Yuqi Nie, Nam H.~Nguyen, Phanwadee Sinthong, and Jayant Kalagnanam.
\newblock A time series is worth 64 words: Long-term forecasting with transformers.
\newblock In \emph{International Conference on Learning Representations}, 2023.

\bibitem[Oh et~al.(2015)Oh, Guo, Lee, Lewis, and Singh]{oh2015action}
Junhyuk Oh, Xiaoxiao Guo, Honglak Lee, Richard~L Lewis, and Satinder Singh.
\newblock Action-conditional video prediction using deep networks in {Atari} games.
\newblock \emph{Advances in neural information processing systems}, 28, 2015.

\bibitem[Oreshkin et~al.(2019)Oreshkin, Carpov, Chapados, and Bengio]{oreshkin2019n}
Boris~N Oreshkin, Dmitri Carpov, Nicolas Chapados, and Yoshua Bengio.
\newblock {N-BEATS}: Neural basis expansion analysis for interpretable time series forecasting.
\newblock \emph{arXiv preprint arXiv:1905.10437}, 2019.

\bibitem[Pascanu et~al.(2013)Pascanu, Mikolov, and Bengio]{pascanu2013difficulty}
Razvan Pascanu, Tomas Mikolov, and Yoshua Bengio.
\newblock On the difficulty of training recurrent neural networks.
\newblock pp.\  III--1310, 2013.

\bibitem[Porter \& Duff(1984)Porter and Duff]{porter1984compositing}
Thomas Porter and Tom Duff.
\newblock Compositing digital images.
\newblock In \emph{Proceedings of the 11th annual conference on Computer graphics and interactive techniques}, pp.\  253--259, 1984.

\bibitem[Razavi et~al.(2019)Razavi, Van~den Oord, and Vinyals]{razavi2019vqvae2}
Ali Razavi, Aaron Van~den Oord, and Oriol Vinyals.
\newblock Generating diverse high-fidelity images with {VQ-VAE-2}.
\newblock In \emph{Advances in neural information processing systems}, 2019.

\bibitem[Robine et~al.(2023)Robine, H{\"o}ftmann, Uelwer, and Harmeling]{robine2023transformer}
Jan Robine, Marc H{\"o}ftmann, Tobias Uelwer, and Stefan Harmeling.
\newblock Transformer-based world models are happy with 100k interactions.
\newblock \emph{arXiv preprint arXiv:2303.07109}, 2023.

\bibitem[Robinson \& Fallside(1989)Robinson and Fallside]{robinson1989dynamic}
Tony Robinson and Frank Fallside.
\newblock Dynamic reinforcement driven error propagation networks with application to game playing.
\newblock In \emph{Proceedings of the Annual Meeting of the Cognitive Science Society}, volume~11, 1989.

\bibitem[Salinas et~al.(2020)Salinas, Flunkert, Gasthaus, and Januschowski]{salinas2020deepar}
David Salinas, Valentin Flunkert, Jan Gasthaus, and Tim Januschowski.
\newblock {DeepAR}: Probabilistic forecasting with autoregressive recurrent networks.
\newblock \emph{International journal of forecasting}, 36\penalty0 (3):\penalty0 1181--1191, 2020.

\bibitem[Samsami et~al.(2024)Samsami, Zholus, Rajendran, and Chandar]{samsami2024mastering}
Mohammad~Reza Samsami, Artem Zholus, Janarthanan Rajendran, and Sarath Chandar.
\newblock Mastering memory tasks with world models.
\newblock \emph{arXiv preprint arXiv:2403.04253}, 2024.

\bibitem[Sara et~al.(2019)Sara, Akter, and Uddin]{sara2019image}
Umme Sara, Morium Akter, and Mohammad~Shorif Uddin.
\newblock Image quality assessment through {FSIM, SSIM, MSE and PSNR}—a comparative study.
\newblock \emph{Journal of Computer and Communications}, 7\penalty0 (3):\penalty0 8--18, 2019.

\bibitem[Schlag et~al.(2021)Schlag, Irie, and Schmidhuber]{schlag2021linear}
Imanol Schlag, Kazuki Irie, and J{\"u}rgen Schmidhuber.
\newblock Linear transformers are secretly fast weight programmers.
\newblock In \emph{International Conference on Machine Learning}, pp.\  9355--9366. PMLR, 2021.

\bibitem[Schmeckpeper et~al.(2020)Schmeckpeper, Xie, Rybkin, Tian, Daniilidis, Levine, and Finn]{schmeckpeper2020learning}
Karl Schmeckpeper, Annie Xie, Oleh Rybkin, Stephen Tian, Kostas Daniilidis, Sergey Levine, and Chelsea Finn.
\newblock Learning predictive models from observation and interaction.
\newblock In \emph{European Conference on Computer Vision}, pp.\  708--725. Springer, 2020.

\bibitem[Schmidhuber(1990{\natexlab{a}})]{schmidhuber1990line}
J{\"u}rgen Schmidhuber.
\newblock An on-line algorithm for dynamic reinforcement learning and planning in reactive environments.
\newblock In \emph{1990 IJCNN international joint conference on neural networks}, pp.\  253--258. IEEE, 1990{\natexlab{a}}.

\bibitem[Schmidhuber(1990{\natexlab{b}})]{schmidhuber1990making}
J{\"u}rgen Schmidhuber.
\newblock Making the world differentiable: On using fully recurrent self-supervised neural networks for dynamic reinforcement learning and planning in non-stationary environments.
\newblock \emph{Institut f{\"u}r Informatik, Technische Universit{\"a}t M{\"u}nchen. Technical Report FKI-126}, 90, 1990{\natexlab{b}}.

\bibitem[Schmidhuber(1991{\natexlab{a}})]{schmidhuber1990reinforcement}
J{\"u}rgen Schmidhuber.
\newblock Reinforcement learning in {Markovian} and {non-Markovian} environments.
\newblock In D.~S. Lippman, J.~E. Moody, and D.~S. Touretzky (eds.), \emph{Advances in Neural Information Processing Systems 3 (NIPS 3)}, pp.\  500--506. Morgan Kaufmann, 1991{\natexlab{a}}.

\bibitem[Schmidhuber(1991{\natexlab{b}})]{schmidhuber1991possibility}
J{\"u}rgen Schmidhuber.
\newblock A possibility for implementing curiosity and boredom in model-building neural controllers.
\newblock pp.\  222--227, 1991{\natexlab{b}}.

\bibitem[Schmidhuber(1992{\natexlab{a}})]{schmidhuber1992fwp}
J{\"u}rgen Schmidhuber.
\newblock Learning to control fast-weight memories: An alternative to dynamic recurrent networks.
\newblock \emph{Neural Computation}, 4\penalty0 (1):\penalty0 131--139, 1992{\natexlab{a}}.

\bibitem[Schmidhuber(1992{\natexlab{b}})]{schmidhuber1992learning}
J{\"u}rgen Schmidhuber.
\newblock Learning complex, extended sequences using the principle of history compression.
\newblock \emph{Neural computation}, 4\penalty0 (2):\penalty0 234--242, 1992{\natexlab{b}}.

\bibitem[Schmidhuber(1992{\natexlab{c}})]{schmidhuber1992pm}
J{\"u}rgen Schmidhuber.
\newblock Learning factorial codes by predictability minimization.
\newblock \emph{Neural computation}, 4\penalty0 (6):\penalty0 863--879, 1992{\natexlab{c}}.

\bibitem[Schmidhuber(2003)]{schmidhuber2003exploring}
J{\"u}rgen Schmidhuber.
\newblock Exploring the predictable.
\newblock In \emph{Advances in evolutionary computing: theory and applications}, pp.\  579--612. Springer, 2003.

\bibitem[Schmidhuber(2015)]{schmidhuber2015learning}
J{\"u}rgen Schmidhuber.
\newblock On learning to think: Algorithmic information theory for novel combinations of reinforcement learning controllers and recurrent neural world models.
\newblock \emph{arXiv preprint arXiv:1511.09249}, 2015.

\bibitem[Schmidhuber et~al.(2007)Schmidhuber, Wierstra, Gagliolo, and Gomez]{schmidhuber2007training}
J{\"u}rgen Schmidhuber, Daan Wierstra, Matteo Gagliolo, and Faustino Gomez.
\newblock Training recurrent networks by {EVOLINO}.
\newblock \emph{Neural computation}, 19\penalty0 (3):\penalty0 757--779, 2007.

\bibitem[Schrittwieser et~al.(2020)Schrittwieser, Antonoglou, Hubert, Simonyan, Sifre, Schmitt, Guez, Lockhart, Hassabis, Graepel, et~al.]{schrittwieser2020mastering}
Julian Schrittwieser, Ioannis Antonoglou, Thomas Hubert, Karen Simonyan, Laurent Sifre, Simon Schmitt, Arthur Guez, Edward Lockhart, Demis Hassabis, Thore Graepel, et~al.
\newblock Mastering {Atari}, go, chess and shogi by planning with a learned model.
\newblock \emph{Nature}, 588\penalty0 (7839):\penalty0 604--609, 2020.

\bibitem[Shi et~al.(2024)Shi, Xia, Jin, Wang, Zhao, Xia, Xiao, and Yang]{shi2024vmambair}
Yuan Shi, Bin Xia, Xiaoyu Jin, Xing Wang, Tianyu Zhao, Xin Xia, Xuefeng Xiao, and Wenming Yang.
\newblock {VmambaIR}: Visual state space model for image restoration.
\newblock \emph{arXiv preprint arXiv:2403.11423}, 2024.

\bibitem[Silver et~al.(2017)Silver, Hasselt, Hessel, Schaul, Guez, Harley, Dulac-Arnold, Reichert, Rabinowitz, Barreto, et~al.]{silver2017predictron}
David Silver, Hado Hasselt, Matteo Hessel, Tom Schaul, Arthur Guez, Tim Harley, Gabriel Dulac-Arnold, David Reichert, Neil Rabinowitz, Andre Barreto, et~al.
\newblock The {Predictron}: End-to-end learning and planning.
\newblock In \emph{International Conference on Machine Learning}, pp.\  3191--3199. PMLR, 2017.

\bibitem[Smith et~al.(2023)Smith, Warrington, and Linderman]{smith2022simplified}
Jimmy~TH Smith, Andrew Warrington, and Scott~W Linderman.
\newblock Simplified state space layers for sequence modeling.
\newblock \emph{International Conference on Learning Representations}, 2023.

\bibitem[Srivastava et~al.(2015)Srivastava, Greff, and Schmidhuber]{srivastava2015training}
Rupesh~K Srivastava, Klaus Greff, and J{\"u}rgen Schmidhuber.
\newblock Training very deep networks.
\newblock \emph{Advances in neural information processing systems}, 28, 2015.

\bibitem[Stani{\'c} et~al.(2023)Stani{\'c}, Tang, Ha, and Schmidhuber]{stanic2023learning}
Aleksandar Stani{\'c}, Yujin Tang, David Ha, and J{\"u}rgen Schmidhuber.
\newblock Learning to generalize with object-centric agents in the open world survival game crafter.
\newblock \emph{IEEE Transactions on Games}, 2023.

\bibitem[Tallec \& Ollivier(2018)Tallec and Ollivier]{tallec2018can}
Corentin Tallec and Yann Ollivier.
\newblock Can recurrent neural networks warp time?
\newblock \emph{arXiv preprint arXiv:1804.11188}, 2018.

\bibitem[Tipping \& Bishop(1999)Tipping and Bishop]{tipping1999probabilistic}
Michael~E Tipping and Christopher~M Bishop.
\newblock Probabilistic principal component analysis.
\newblock \emph{Journal of the Royal Statistical Society Series B: Statistical Methodology}, 61\penalty0 (3):\penalty0 611--622, 1999.

\bibitem[Tishby et~al.(2000)Tishby, Pereira, and Bialek]{tishby2000information}
Naftali Tishby, Fernando~C Pereira, and William Bialek.
\newblock The information bottleneck method.
\newblock \emph{arXiv preprint physics/0004057}, 2000.

\bibitem[Vaswani et~al.(2017)Vaswani, Shazeer, Parmar, Uszkoreit, Jones, Gomez, Kaiser, and Polosukhin]{vaswani2017attention}
Ashish Vaswani, Noam Shazeer, Niki Parmar, Jakob Uszkoreit, Llion Jones, Aidan~N Gomez, {\L}ukasz Kaiser, and Illia Polosukhin.
\newblock Attention is all you need.
\newblock \emph{Advances in Neural Information Processing Systems}, 2017.

\bibitem[Wang et~al.(2023)Wang, Zhu, Wang, Yu, Liu, Omar, and Hamid]{wang2023selective}
Jue Wang, Wentao Zhu, Pichao Wang, Xiang Yu, Linda Liu, Mohamed Omar, and Raffay Hamid.
\newblock Selective structured state-spaces for long-form video understanding.
\newblock In \emph{Proceedings of the IEEE/CVF Conference on Computer Vision and Pattern Recognition}, pp.\  6387--6397, 2023.

\bibitem[Wang et~al.(2024{\natexlab{a}})Wang, Wu, Shi, Hu, Luo, Ma, Zhang, and ZHOU]{wang2023timemixer}
Shiyu Wang, Haixu Wu, Xiaoming Shi, Tengge Hu, Huakun Luo, Lintao Ma, James~Y Zhang, and JUN ZHOU.
\newblock {TimeMixer}: Decomposable multiscale mixing for time series forecasting.
\newblock In \emph{International Conference on Learning Representations (ICLR)}, 2024{\natexlab{a}}.

\bibitem[Wang et~al.(2024{\natexlab{b}})Wang, Wang, Ding, Li, Wu, Rong, Kong, Huang, Li, Yang, et~al.]{wang2024state}
Xiao Wang, Shiao Wang, Yuhe Ding, Yuehang Li, Wentao Wu, Yao Rong, Weizhe Kong, Ju~Huang, Shihao Li, Haoxiang Yang, et~al.
\newblock State space model for new-generation network alternative to transformers: A survey.
\newblock \emph{arXiv preprint arXiv:2404.09516}, 2024{\natexlab{b}}.

\bibitem[Wang et~al.(2017)Wang, Long, Wang, Gao, and Yu]{wang2017predrnn}
Yunbo Wang, Mingsheng Long, Jianmin Wang, Zhifeng Gao, and Philip~S Yu.
\newblock {PredRNN}: Recurrent neural networks for predictive learning using spatiotemporal {LSTMs}.
\newblock In \emph{Advances in neural information processing systems}, 2017.

\bibitem[Wang et~al.(2024{\natexlab{c}})Wang, Kong, Feng, Wang, Zhao, Wang, and Zhang]{wang2024smamba}
Zihan Wang, Fanheng Kong, Shi Feng, Ming Wang, Han Zhao, Daling Wang, and Yifei Zhang.
\newblock Is {Mamba} effective for time series forecasting?
\newblock \emph{arXiv preprint arXiv:2403.11144}, 2024{\natexlab{c}}.

\bibitem[Watters et~al.(2017)Watters, Zoran, Weber, Battaglia, Pascanu, and Tacchetti]{watters2017visual}
Nicholas Watters, Daniel Zoran, Theophane Weber, Peter Battaglia, Razvan Pascanu, and Andrea Tacchetti.
\newblock {Visual Interaction Networks}: Learning a physics simulator from video.
\newblock \emph{Advances in neural information processing systems}, 30, 2017.

\bibitem[Watters et~al.(2019)Watters, Matthey, Burgess, and Lerchner]{watters2019spatial}
Nicholas Watters, Loic Matthey, Christopher~P Burgess, and Alexander Lerchner.
\newblock {Spatial Broadcast Decoder}: A simple architecture for learning disentangled representations in vaes.
\newblock \emph{arXiv preprint arXiv:1901.07017}, 2019.

\bibitem[Werbos(1987)]{werbos1987learning}
Paul~J Werbos.
\newblock Learning how the world works: Specifications for predictive networks in robots and brains.
\newblock In \emph{Proceedings of IEEE International Conference on Systems, Man and Cybernetics, NY}, 1987.

\bibitem[Werbos(1989)]{werbos1989neural}
Paul~J Werbos.
\newblock Neural networks for control and system identification.
\newblock In \emph{Proceedings of the 28th IEEE Conference on Decision and Control,}, pp.\  260--265. IEEE, 1989.

\bibitem[Williams \& Titsias(2004)Williams and Titsias]{williams2004greedy}
Christopher K~I Williams and Michalis~K Titsias.
\newblock Greedy learning of multiple objects in images using robust statistics and factorial learning.
\newblock \emph{Neural Computation}, 16\penalty0 (5):\penalty0 1039--1062, 2004.

\bibitem[Wu et~al.(2021)Wu, Xu, Wang, and Long]{wu2021autoformer}
Haixu Wu, Jiehui Xu, Jianmin Wang, and Mingsheng Long.
\newblock {Autoformer}: Decomposition transformers with auto-correlation for long-term series forecasting.
\newblock In \emph{Advances in Neural Information Processing Systems}, 2021.

\bibitem[Wu et~al.(2023{\natexlab{a}})Wu, Hu, Liu, Zhou, Wang, and Long]{wu2022timesnet}
Haixu Wu, Tengge Hu, Yong Liu, Hang Zhou, Jianmin Wang, and Mingsheng Long.
\newblock {TimesNet}: Temporal 2d-variation modeling for general time series analysis.
\newblock In \emph{International Conference on Learning Representations}, 2023{\natexlab{a}}.

\bibitem[Wu et~al.(2023{\natexlab{b}})Wu, Dvornik, Greff, Kipf, and Garg]{wu2022slotformer}
Ziyi Wu, Nikita Dvornik, Klaus Greff, Thomas Kipf, and Animesh Garg.
\newblock {SlotFormer}: Unsupervised visual dynamics simulation with object-centric models.
\newblock In \emph{International Conference on Learning Representations}, 2023{\natexlab{b}}.

\bibitem[Wu et~al.(2019)Wu, Pan, Long, Jiang, and Zhang]{ijcai2019p264}
Zonghan Wu, Shirui Pan, Guodong Long, Jing Jiang, and Chengqi Zhang.
\newblock Graph wavenet for deep spatial-temporal graph modeling.
\newblock In \emph{Proceedings of the Twenty-Eighth International Joint Conference on Artificial Intelligence, {IJCAI-19}}, pp.\  1907--1913. International Joint Conferences on Artificial Intelligence Organization, 7 2019.
\newblock \doi{10.24963/ijcai.2019/264}.
\newblock URL \url{https://doi.org/10.24963/ijcai.2019/264}.

\bibitem[Wu et~al.(2020{\natexlab{a}})Wu, Pan, Chen, Long, Zhang, and Philip]{wu2020gnnsurvey}
Zonghan Wu, Shirui Pan, Fengwen Chen, Guodong Long, Chengqi Zhang, and S~Yu Philip.
\newblock A comprehensive survey on graph neural networks.
\newblock \emph{IEEE transactions on neural networks and learning systems}, 32\penalty0 (1):\penalty0 4--24, 2020{\natexlab{a}}.

\bibitem[Wu et~al.(2020{\natexlab{b}})Wu, Pan, Long, Jiang, Chang, and Zhang]{wu2020connecting}
Zonghan Wu, Shirui Pan, Guodong Long, Jing Jiang, Xiaojun Chang, and Chengqi Zhang.
\newblock Connecting the dots: Multivariate time series forecasting with graph neural networks.
\newblock In \emph{Proceedings of the 26th ACM SIGKDD international conference on knowledge discovery \& data mining}, pp.\  753--763, 2020{\natexlab{b}}.

\bibitem[Yan et~al.(2024)Yan, Gu, and Rush]{yan2024diffusion}
Jing~Nathan Yan, Jiatao Gu, and Alexander~M Rush.
\newblock Diffusion models without attention.
\newblock In \emph{Proceedings of the IEEE/CVF Conference on Computer Vision and Pattern Recognition}, pp.\  8239--8249, 2024.

\bibitem[Yang et~al.(2024{\natexlab{a}})Yang, Ma, Yao, Zhong, Zhang, and Wang]{yang2024remamber}
Yuhuan Yang, Chaofan Ma, Jiangchao Yao, Zhun Zhong, Ya~Zhang, and Yanfeng Wang.
\newblock Remamber: Referring image segmentation with mamba twister.
\newblock \emph{arXiv preprint arXiv:2403.17839}, 2024{\natexlab{a}}.

\bibitem[Yang et~al.(2024{\natexlab{b}})Yang, Mitra, Kwon, and Yu]{yang2024clinicalmamba}
Zhichao Yang, Avijit Mitra, Sunjae Kwon, and Hong Yu.
\newblock {ClinicalMamba}: A generative clinical language model on longitudinal clinical notes.
\newblock \emph{arXiv preprint arXiv:2403.05795}, 2024{\natexlab{b}}.

\bibitem[Yi et~al.(2020)Yi, Gan, Li, Kohli, Wu, Torralba, and Tenenbaum]{yiclevrer}
Kexin Yi, Chuang Gan, Yunzhu Li, Pushmeet Kohli, Jiajun Wu, Antonio Torralba, and Joshua~B Tenenbaum.
\newblock {CLEVRER}: Collision events for video representation and reasoning.
\newblock In \emph{International Conference on Learning Representations}, 2020.

\bibitem[Yu et~al.(2018)Yu, Yin, and Zhu]{ijcai2018p505}
Bing Yu, Haoteng Yin, and Zhanxing Zhu.
\newblock Spatio-temporal graph convolutional networks: A deep learning framework for traffic forecasting.
\newblock In \emph{Proceedings of the Twenty-Seventh International Joint Conference on Artificial Intelligence, {IJCAI-18}}, pp.\  3634--3640. International Joint Conferences on Artificial Intelligence Organization, 7 2018.
\newblock \doi{10.24963/ijcai.2018/505}.
\newblock URL \url{https://doi.org/10.24963/ijcai.2018/505}.

\bibitem[Yuille \& Kersten(2006)Yuille and Kersten]{yuille2006vision}
Alan Yuille and Daniel Kersten.
\newblock Vision as {Bayesian} inference: analysis by synthesis?
\newblock \emph{Trends in cognitive sciences}, 2006.

\bibitem[Zeng et~al.(2023)Zeng, Chen, Zhang, and Xu]{zeng2023dlinear}
Ailing Zeng, Muxi Chen, Lei Zhang, and Qiang Xu.
\newblock Are transformers effective for time series forecasting?
\newblock In \emph{Proceedings of the AAAI conference on artificial intelligence}, volume~37, pp.\  11121--11128, 2023.

\bibitem[Zhang et~al.(2022)Zhang, Roller, Goyal, Artetxe, Chen, Chen, Dewan, Diab, Li, Lin, et~al.]{zhang2022opt}
Susan Zhang, Stephen Roller, Naman Goyal, Mikel Artetxe, Moya Chen, Shuohui Chen, Christopher Dewan, Mona Diab, Xian Li, Xi~Victoria Lin, et~al.
\newblock {OPT}: Open pre-trained transformer language models.
\newblock \emph{arXiv preprint arXiv:2205.01068}, 2022.

\bibitem[Zhang \& Yan(2023)Zhang and Yan]{zhang2023crossformer}
Yunhao Zhang and Junchi Yan.
\newblock {Crossformer}: Transformer utilizing cross-dimension dependency for multivariate time series forecasting.
\newblock In \emph{International Conference on Learning Representations}, 2023.

\bibitem[Zheng et~al.(2020)Zheng, Fan, Wang, and Qi]{zheng2020gman}
Chuanpan Zheng, Xiaoliang Fan, Cheng Wang, and Jianzhong Qi.
\newblock Gman: A graph multi-attention network for traffic prediction.
\newblock In \emph{Proceedings of the AAAI conference on artificial intelligence}, volume~34, pp.\  1234--1241, 2020.

\bibitem[Zhou et~al.(2022)Zhou, Ma, Wen, Wang, Sun, and Jin]{zhou2022fedformer}
Tian Zhou, Ziqing Ma, Qingsong Wen, Xue Wang, Liang Sun, and Rong Jin.
\newblock {FEDformer}: Frequency enhanced decomposed transformer for long-term series forecasting.
\newblock In \emph{International Conference on Machine Learning}, 2022.

\end{thebibliography}
